\documentclass[10pt,twocolumn,letterpaper]{article}

\usepackage[colorinlistoftodos,prependcaption,textsize=tiny, textwidth=2cm]{todonotes}

\usepackage[pagenumbers]{cvpr}      %

\definecolor{cvprblue}{rgb}{0.21,0.49,0.74}

\usepackage{graphicx}
\usepackage{wrapfig,booktabs}
\usepackage{mathtools,xparse}
\usepackage{multirow}
\usepackage{multicol}
\usepackage{color}
\usepackage{colortbl}
\usepackage{pifont}

\colorlet{darkgreen}{green!65!black}
\colorlet{darkblue}{blue!75!black}
\colorlet{darkred}{red!80!black}
\definecolor{lightblue}{HTML}{0071bc}
\definecolor{lightgreen}{HTML}{39b54a}

\newcolumntype{g}{>{\columncolor{tbgray}}c}

\definecolor{red}{rgb}{0.8, 0.0, 0.0}
\definecolor{green}{rgb}{0.0, 0.5, 0.0}
\definecolor{tbgray}{gray}{.92}

\usepackage[pagebackref,breaklinks,colorlinks,allcolors=cvprblue]{hyperref}

\usepackage{xargs}   
\newcommandx{\zkn}[2][1=]{\todo[linecolor=red,backgroundcolor=red!25,bordercolor=blue,#1]{#2}}

\newcommandx{\shalbe}[2][1=]{\todo[linecolor=red,backgroundcolor=red!25,bordercolor=red,#1]{#2}}

\newcolumntype{L}[1]{>{\raggedright\arraybackslash}p{#1}}
\newcolumntype{C}[1]{>{\centering\arraybackslash}p{#1}}

\definecolor{rowgray}{gray}{0.95}
\def\paperID{9628} %
\def\confName{CVPR}
\def\confYear{2025}

\title{Grounding Descriptions in Images informs Zero-Shot Visual Recognition} 

\author{
\textbf{
Shaunak Halbe\thanks{Correspondence to shalbe9@gatech.edu}\,\,\textsuperscript{1}
\quad Junjiao Tian\textsuperscript{1}
\quad K J Joseph\textsuperscript{2}
\quad James Seale Smith\textsuperscript{1}
}
\\
\textbf{
\quad Katherine Stevo\textsuperscript{1}
\quad Vineeth N Balasubramanian\textsuperscript{3}
\quad Zsolt Kira\textsuperscript{1}
}
\\
\normalsize
\textsuperscript{1}Georgia Institute of Technology
\quad  \textsuperscript{2}Adobe Research
\normalsize
\quad \textsuperscript{3}Indian Institute of Technology, Hyderabad
}

\begin{document}
\maketitle
\begin{abstract}
  Vision-language models (VLMs) like CLIP have been cherished for their ability to perform zero-shot visual recognition on open-vocabulary concepts. This is achieved by selecting the object category whose textual representation bears the highest similarity with the query image. While successful in some domains, this method struggles with identifying fine-grained entities as well as generalizing to unseen concepts that are not captured by the training distribution. Recent works attempt to mitigate these challenges by integrating category descriptions at test time, albeit yielding modest improvements. We attribute these limited gains to a fundamental misalignment between image and description representations, which is rooted in the pretraining structure of CLIP. In this paper, we propose \textit{GRAIN}, a new pretraining strategy aimed at aligning representations at both fine and coarse levels simultaneously. Our approach learns to jointly ground textual descriptions in image regions along with aligning overarching captions with global image representations. To drive this pre-training, we leverage frozen Multimodal Large Language Models (MLLMs)
   to derive large-scale synthetic annotations.  We demonstrate the enhanced zero-shot performance of our model compared to current state-of-the art methods across 11 diverse image classification datasets. 
Additionally, we introduce \textit{Products-2023}, a newly curated, manually labeled dataset featuring novel concepts, and showcase our model's ability to recognize these concepts by benchmarking on it. Significant improvements achieved by our model on other downstream tasks like retrieval further highlight the superior quality of representations learned by our approach. Code available at \url{https://github.com/shaunak27/grain-clip}.

\end{abstract}
\section{Introduction}
\label{sec:intro}

Traditionally, image classification has operated under the closed-set assumption where models are evaluated on a fixed set of classes that were seen during training. However, in the real and open-world, models need to account for test conditions where the number of classes is unknown during training and can include classes that were not seen.Vision-language models (VLMs) like CLIP \cite{radfordlearning} offer a solution in this space,  owing to their \textit{open-vocabulary} nature. These models undergo extensive pretraining on large datasets containing paired image-text data and learn to encode images and texts in a shared latent space where semantically similar representations are mapped closed together. For zero-shot classification, CLIP leverages the names of all classes within the test dataset—referred to as the \textit{vocabulary}—as the candidate set, and determines the most probable image-classname pairing by computing the similarity between their latent representations. This vocabulary of classes is unconstrained, enabling the inclusion of any concept, regardless of its presence in the training set.  This facilitates classification from an \textit{open-set} of concepts. 

Despite this, CLIP's zero-shot capabilities are still limited by a few critical challenges. Firstly, in practice, CLIP often struggles to differentiate between fine-grained categories, a limitation highlighted by its under-performance on Fine-Grained Visual Classification (FGVC) datasets \cite{maji2013finegrained,WahCUB_200_2011}. Secondly, while known for its open-vocabulary potential, it can still perform poorly for some domains not well-represented in the training distribution, especially if the vocabulary used has confounding categories during testing.  %
Using a vocabulary that exceeds the scope of the test dataset significantly diminishes the performance of CLIP even for common datasets like Imagenet~\cite{hu2023open}.  This decline is again largely attributed to CLIP's challenges in differentiating between semantically similar, fine-grained concepts.  Additionally, CLIP's inability to recognize novel concepts, such as \verb|Apple Vision Pro| that were not present during its training phase, further restricts its capability to function as a genuinely open-vocabulary model.

Recent works \cite{menon2023visual,pratt2023does} aim to address these challenges by incorporating extra information in the form of class descriptions generated by Large Language Models (LLMs) at test time. These approaches leverage the ``visual" knowledge embedded in LLMs to augment the textual representations used in zero-shot classification. As an example, the class \verb|French Bulldog| would be expanded to \verb|A French Bulldog, which has small and| \verb|pointy ears.| These methods provide some improvements over standard CLIP models, though they leave room for further advancements.

\looseness=-1 We hypothesize that the reason for achieving limited gains from injecting these descriptions lies in the poor alignment between image and description embeddings learned by CLIP. As a result, we aim to verify this hypothesis and propose a method to overcome these challenges.
Specifically, we posit that the misalignment between images and descriptions stems from CLIP's training structure, which focuses solely on the global objective of matching entire images to their overarching captions, neglecting the rich information that image regions and textual descriptions share with each other. Our observations align with recent research indicating that CLIP tends to overlook fine-grained visual details during pretraining, leading to subpar performance on tasks requiring localization \cite{ranasinghe2023perceptual}, object attributes \cite{yuksekgonul2023visionlanguage}, or physical reasoning \cite{parcalabescu-etal-2022-valse}.

In this work, we propose \textit{GRAIN: Grounding and contrastive alignment of descriptions}, a novel objective for contrastive vision-language pretraining that learns representations more conducive to zero-shot visual recognition. This is  achieved through fine-grained correspondence between image regions and detailed text descriptions. As a first step towards our approach, given that pretraining datasets (Conceptual Captions~\cite{sharma2018conceptual}, LAION~\cite{schuhmann2022laion}, etc.) only contain images with noisy captions but without detailed descriptions, we employ an instruction-tuned Multimodal Large Language Model (MLLM) to generate descriptions and identify salient attributes from the images in these datasets. Following this, we acquire region-level annotations that correspond to these descriptions using an off-the-shelf Open-vocabulary Object Detector (OVD). We then propose a method that learns to jointly ground text descriptions into specific image regions along with aligning  image and caption representations at a global level. This strategy aims to learn representations that encode both coarse-grained (global) and fine-grained (local) information. To achieve this, we introduce a query transformer architecture for encoding images and a text encoder for processing captions and descriptions. The architecture and objectives of our model are specifically crafted to learn object/region-aware image representations that are valuable for zero-shot tasks as we demonstrate in the subsequent sections. Finally, to evaluate our model's ability to recognize novel concepts, we curate and manually label a new image classification dataset, \textit{Products-2023}, and benchmark upon it.

\vspace{4pt}
\noindent To summarize, our main contributions are as follows:
\vspace{4pt}
\begin{itemize}
    \item We hypothesize and show that CLIP pre-training lacks fine-grained aligned representations, leading to poor zero-shot performance in some domains. 
    \item We propose \textit{GRAIN}, a novel pre-training architecture and objective designed to simultaneously learn local and global correspondences, obtained via weak supervision from  Multimodal LLMs and open-vocabulary detectors. 
    \item To drive this pre-training, we introduce an automated annotation engine to source fine-grained supervision signal.
    \item We demonstrate significant gains across a range of tasks, including image classification and retrieval, specifically improve over the state-of-art by up to \textbf{9\%} in absolute top-1 accuracy for zero-shot classification and up to \textbf{25\%} across cross-modal retrieval tasks.
    \item Acknowledging the lack of novel image classification datasets, we collect and manually label a dataset, \textit{Products-2023}, for benchmarking, which we plan to release.
    \item Additionally, we aim to release our pre-trained model weights along with the large-scale annotations to aid future research.
\end{itemize}

\section{Related Works}
\begin{figure}[t]
    \centering
    \centering
    \includegraphics[width=0.5\textwidth,height=0.25\textwidth]{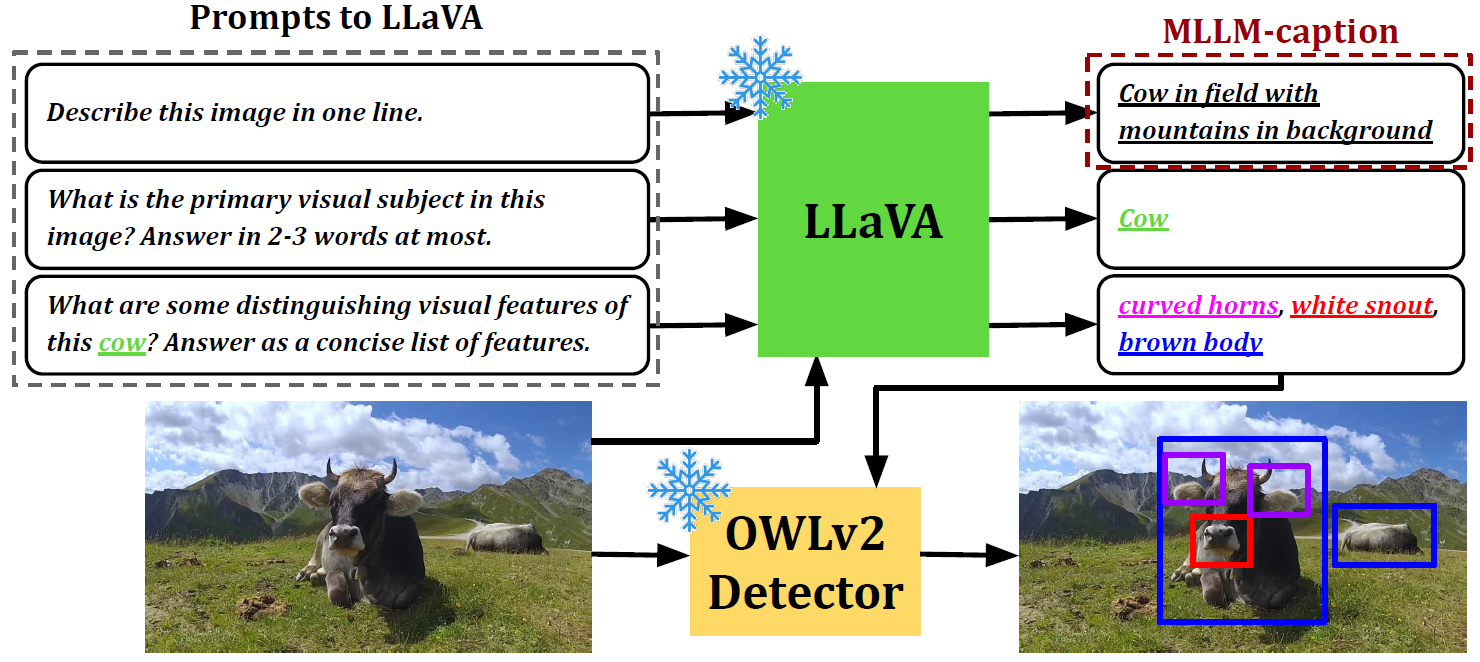}
    \caption{Overview of our two-stage annotation process: (1) prompting LLaVA for image descriptions and (2) acquiring corresponding region annotations from OWLv2.}
    \label{fig:annotation}
\end{figure}
\textbf{Contrastive Language-Image Pretraining.}
\begin{figure*}[t]
    \centering
    \centering
    \includegraphics[width=0.9\textwidth,height=0.4\textwidth]{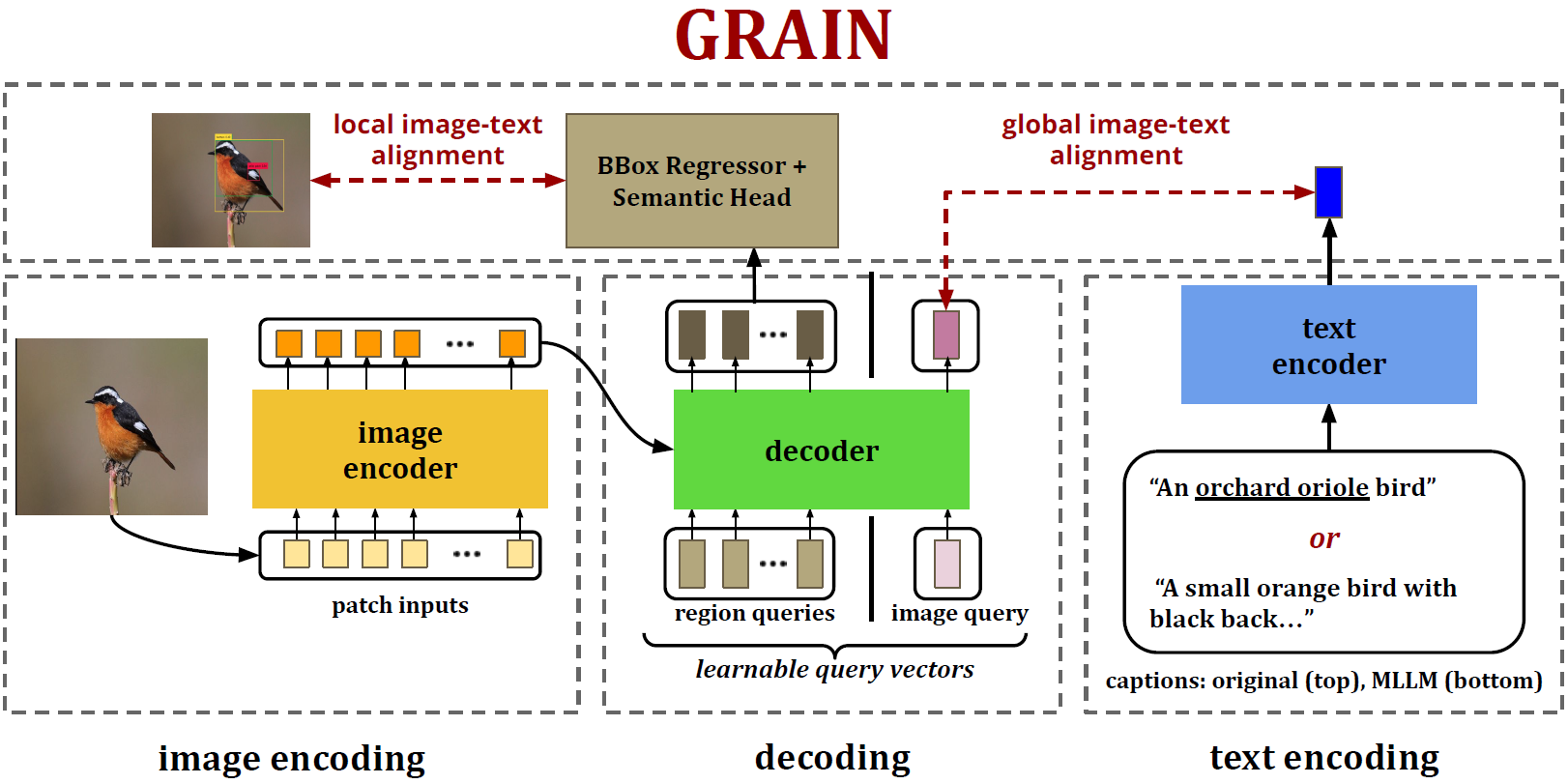}
    \caption{Architecture overview. Our method, GRAIN, aligns image representations to text captions at a global level while localizing salient image regions and aligning them to text descriptions at the local level.}
    \label{fig:method}
    \vspace{-3mm}
\end{figure*}
Follow-up works on CLIP \cite{radford2021learning} and ALIGN \cite{jia2021scaling} focus on improving the quality of learned representations by further introducing self-supervision or cross-modal alignment objectives \cite{mu2021slip,zhai2023sigmoid,9710099}. Relevant to our focus, FILIP \cite{yao2021filip} introduces a cross-modal late interaction mechanism that explores token-wise maximum similarity between image and text tokens to improve fine-grained alignment. Recently, SPARC \cite{bica2024improving} proposes a sparse similarity metric between image patches and text tokens to learn fine-grained representations. While our paper shares motivation with these works, we address the fact that web-based captioning datasets \cite{sharma2018conceptual,schuhmann2022laion} contain noisy captions that lack descriptive information thereby limiting the gains achievable from such elaborate objectives. Instead, we source rich text descriptions and region annotations and design a pre-training objective to learn from them. This allows us to effectively use complementary information at test-time (in the form of LLM-generated descriptions) to recognize fine-grained or novel entities.%

\vspace{4pt}
\textbf{Improving CLIP using Generative Models.}
Recent works have explored the use of LLMs towards improving the downstream performance of CLIP. Menon \etal \cite{menon2023visual} and CuPL \cite{pratt2023does} focus on the task of zero-shot classification, and prompt GPT-3 \cite{brown2020language} at test-time to generate class descriptions. These descriptions are integrated into the classification prompts to achieve gains in terms of accuracy and interpretability. Different from these, LaCLIP \cite{fan2023improving} and VeCLIP \cite{lai2024veclipimprovingcliptraining} use LLMs to rephrase captions from pretraining datasets and observe noticeable gains on downstream tasks by training on these captions. In this paper, we propose to leverage synthetic annotations in the form on image regions and descriptions generated by a MLLM and an open-world detector to drive a novel pretraining strategy.%

\vspace{4pt}

\section{Approach}

\looseness=-1 We propose GRAIN, a novel pretraining approach that simultaneously learns local and global correspondences between image and text representations. Motivated by the observation that CLIP representations lack sufficient fine-grained understanding, we introduce a transformer-based architecture inspired by DETR \cite{carion2020endtoend}, to infuse the rich context from sub-image regions into learned visual representations. Alongside encoding the image into a semantic representation, our model predicts bounding boxes for salient image regions containing discriminative information. These localizations are then aligned with detailed textual descriptions. To supervise this fine-grained objective, we first generate annotations at scale by leveraging Multimodal Large Language Models (MLLMs) and Open-vocabulary Object Detectors (OVDs).  In this section, we first elaborate our automated annotation process and then proceed to discussing our architecture and training methodology.

\subsection{Weak Supervision from MLLMs and OVDs}
We utilize the 3M and 12M versions of the Conceptual Captions~\cite{sharma2018conceptual} (CC3M, CC12M) dataset to train our model. These datasets contain images sourced from the internet, each paired with corresponding alt-texts (or captions). Our approach requires region-level supervision that is not provided by any existing dataset at scale.
Specifically, we find that the captions associated with these images are often noisy, lack detail and may not fully capture the dense visual context. To learn fine-grained correspondence between the two modalities, we propose focusing on regions within the image and their descriptions in text as supervision for training our model.  For generating descriptions and locating their corresponding regions, we leverage an instruction-tuned Multimodal Large Language Model,  LLaVA\cite{liu2024visual}. We select LLaVA for its superior captioning capabilities and accessibility due to its openness;  however, our approach is fundamentally compatible with any multimodal LLM.
For our annotation purposes, we select the LLaVA v1.6 model which integrates a pretrained Vision Transformer Large (ViT-L) \cite{dosovitskiy2020image} as the visual encoder with the Vicuna-13B LLM \cite{vicuna2023}. It is worth noting that we only use LLaVA to describe regions/components of the image at a high level and not pinpoint specific fine-grained categories. A common problem with instruction-tuned models like LLaVA is their tendency to hallucinate, which causes the model to output sentences that are not well-grounded in the image. To address this, we propose a two-stage  approach, as illustrated in Figure \ref{fig:annotation}, to elicit accurate descriptions from LLaVA while minimizing hallucination.
\begin{figure}[t]
    \centering
    \centering
    \includegraphics[width=0.48\textwidth,height=0.2\textwidth]{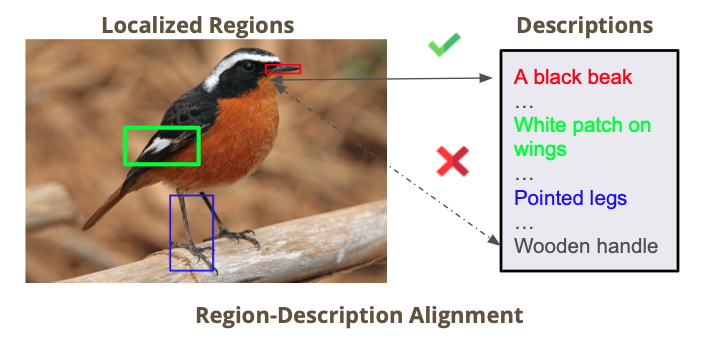}
    \caption{Contrastively align predicted regions with descriptions.}
    \label{fig:alignment}
    \vspace{-5mm}
\end{figure}

Specifically, the two-stage prompting approach is as follows: in the first stage, we ask LLaVA to identify the primary visual subject in the image using a simple, fixed prompt: ``\textit{What is the primary visual subject in this image? Answer in 2-3 words at most.}'' By doing this for every image, we collect the main focus of each image. The generations from this prompt typically capture the prominent object, scene, or concept at a high level.
Next, we construct specific prompts for each image by asking LLaVA to describe the identified subject: ``\textit{What are some distinguishing visual features of this \{subject\}? Answer as a concise list of features}''. We observe that the generations from this two-stage pipeline are more faithful to the visual context and less susceptible to hallucinations.%
We present a qualitative analysis on this in the Appendix. This procedure provides us with a list of descriptions for each image.
Additionally, we ask LLaVA to generate a short one-line description the image by prompting it with ``\textit{Describe this image in one line}''. This description gives a high-level overview of the visual context, and it is utilized as text-level data augmentation during training. From this point forward, we refer to this description as the \textit{MLLM-caption}, and the one from the pretraining dataset as the \textit{original caption}.

Next, we are tasked with localizing these generated descriptions within the image to obtain the necessary supervision for training our grounding module. We leverage the OWLv2 Open-vocabulary Detector~\cite{minderer2024scaling} to localize these descriptions within the image. For each description, we filter out the core attribute being referred to and pass it to the open-world detector for localization. The detector generates several candidate proposals, from which we select detections based on a confidence threshold value. We set this threshold to a relatively high value to ensure high-quality detections. Subsequently, we eliminate redundant bounding box predictions using non-maximum suppression, retaining only the box with the highest confidence score for each region and discarding others with significant overlap.

This procedure enables us to acquire descriptions, bounding boxes, and MLLM captions, which are subsequently utilized to train our model, as detailed in the upcoming section.  To our knowledge, we are the first to obtain such fine-grained annotations on a large scale. The overall annotation process took around 600 GPU hours for CC3M and $\small \sim$ 2200 GPU hours for CC12M using NVIDIA A40s. We aim to release this dataset to benefit future research in this direction.

\begin{table*}[t]
\begin{center}
\caption{\small{Zero-shot transfer evaluation of different models. We highlight the best performance of each setting in {\bf{bold}}. We see that GRAIN improves performance under both pretraining datasets, outperforming CLIP by up to \textbf{9\%} in absolute top-1 accuracy. CLIP* is a version of CLIP with the same number of parameters as our method for fair comparison.
}}
\vspace*{1.5mm}
\label{table:zeroshot-main}
\resizebox{1.0\textwidth}{!}{
\begin{tabular}{c@{\hspace{1.9em}}c@{\hspace{0.5em}}|@{\hspace{0.5em}}c@{\hspace{0.5em}}c@{\hspace{0.5em}}c@{\hspace{0.5em}}c@{\hspace{0.5em}}c@{\hspace{0.5em}}c@{\hspace{0.5em}}c@{\hspace{0.5em}}c@{\hspace{0.5em}}c@{\hspace{0.5em}}c@{\hspace{0.5em}}c@{\hspace{0.5em}}c@{\hspace{0.5em}}|@{\hspace{0.5em}}c@{\hspace{0.5em}}}

\toprule[1.2pt]
\bf Data&\bf Model&
\rotatebox[origin=lb]{90}{\smash{\small CIFAR-10}} & \rotatebox[origin=lb]{90}{\smash{\small CIFAR-100}} &
\rotatebox[origin=lb]{90}{\smash{\small SUN397}} & \rotatebox[origin=lb]{90}{\smash{\small Cars}} & \rotatebox[origin=lb]{90}{\smash{\small DTD}} & \rotatebox[origin=lb]{90}{\smash{\small Pets}} & \rotatebox[origin=lb]{90}{\smash{\small Caltech-101}} &
\rotatebox[origin=lb]{90}{\smash{\small Flowers}} & \rotatebox[origin=lb]{90}{\smash{\small CUB}} &  \rotatebox[origin=lb]{90}{\smash{\small Places365}} & \rotatebox[origin=lb]{90}{\smash{\small Food101}}  & \rotatebox[origin=lb]{90}{\smash{\small \bf Average}} & \rotatebox[origin=lb]{90}{\smash{\small \bf ImageNet}} \\
\midrule
 \multirow{1}{1.8em}{\rotatebox[origin=c]{0}{\small }}  &  \small \textcolor{gray}{LLaVA + CLIP}&  \textcolor{gray}{89.69} &  \textcolor{gray}{57.72} & \textcolor{gray}{55.24} & \textcolor{gray}{15.90} & \textcolor{gray}{35.37} & \textcolor{gray}{47.16} & \textcolor{gray}{75.03} & \textcolor{gray}{24.69} & \textcolor{gray}{6.22} &  \textcolor{gray}{29.43}  & \textcolor{gray}{52.80} & \textcolor{gray}{44.48} & \textcolor{gray}{35.20}\\    
\midrule
\multirow{7}{1.5em}{\rotatebox[origin=c]{0}{\small CC3M}}  & \small CLIP\cite{radford2021learning} &  48.86 & 18.70 &  28.44 & 0.68 & 9.23 & 6.94 & 41.02 & 8.48 & 2.51 &  17.85 & 8.73 & 17.40 & 14.01 \\

& \small Menon\&Vondrick\cite{menon2023visual} &  49.35 & 17.93 &  29.74 & 0.60 & 10.43 & 7.05 & 43.89 & 7.67 & 2.84 &  19.12  & 9.64 & 18.02 & 14.12 \\
                                                    & \small CuPL\cite{pratt2023does} &  50.16 & 18.98 & 29.66  &  0.71 & 9.89 & 8.22 & 43.95 & 8.84 & 2.91 &  19.73  & 10.51 & 18.51 & 14.14 \\
                                                    \cmidrule{2-15}
                                                    & \small CLIP* &  46.99 & 18.49 & 29.76 & 0.52 & 8.40 & 6.62 & 42.56 & 8.29 & 3.36 &  18.70  & 10.01 & 17.62 & 14.04 \\
                                                    & \small CLIP* + Menon\&Vondrick\cite{menon2023visual} &  49.37 & 17.98 &  29.94 & 0.62 & 10.55 & 7.14 & 44.02 & 8.38 & 3.51 &  19.23  & 10.24 & 18.27 & 14.16 \\
                                                    & \small CLIP* + CuPL\cite{pratt2023does} &  50.24 & 18.86 & 30.12  &  0.74 & 10.14 & 8.06 & 43.78 & 8.95 & 3.32 &  19.56 & 10.77 & 18.59 & 14.14 \\
                                                    \cmidrule{2-15}
                                                    & \small GRAIN (Ours) & \bf 65.86 & \bf 35.20 &  \bf 38.07 & \bf 1.34 & \bf 17.24 & \bf 14.15 & \bf 65.20 & \bf 13.24 & \bf 5.47 & \bf 24.96 & \bf 16.18 & \bf 27.00 & \bf 23.34 \\
\midrule
\multirow{7}{1.8em}{\rotatebox[origin=c]{0}{\small CC12M}}   & \small CLIP \cite{radford2021learning} &  71.24 &  36.66 & 48.84 & 4.57 & 19.28 & 42.06 & 70.09 & 20.51 & 7.63 & 31.84 & 40.94 & 35.79 & 34.66 \\
& \small Menon\&Vondrick \cite{menon2023visual} &  72.68 & 37.08 &  48.59 & 5.12 & 18.45 & 41.38 & 72.29 & 21.15 & 8.27 &  31.36  & 41.20 & 36.14 & 34.32 \\
                                                    & \small CuPL \cite{pratt2023does} & 72.85 & 37.37 &  49.06 & 4.88 & 18.71 & 41.58 & 71.17 & 22.82 & 7.94 &  30.28  & 40.89 & 36.15 & 34.65 \\
                                                    \cmidrule{2-15}
& \small CLIP* &  70.07 & 35.63 &  50.42 & 4.31 & 18.35 & 39.40 & 74.24 & 21.04 & 7.96 & 32.03 & 41.36 & 35.89  & 33.51 \\  & \small CLIP* + Menon\&Vondrick \cite{menon2023visual} &  72.74 & 37.44 &  51.20 & 5.31 & 18.47 & 41.74 & 74.44 & 21.22 & 8.32 &  32.72  & 41.92 & 36.87 & 34.50 \\
                                                    & \small CLIP* + CuPL \cite{pratt2023does} & 72.77 & 37.85 &  51.08 & 5.12 & 18.98 & 41.14 & 74.22 & 22.68 & 8.05 &  32.34  & 41.65 & 36.90 & 34.77 \\
                                                    \cmidrule{2-15}
                                                    & \small GRAIN (Ours) & \bf 81.40 &  \bf 46.23 & \bf 55.26 & \bf 8.42 & \bf 25.68 & \bf 48.76 & \bf 81.49 & \bf 26.27 & \bf 10.28 &  \bf 36.76  & \bf 45.39 & \bf 42.36 & \bf 41.46 \\

\bottomrule[1.2pt]
\end{tabular}}
\end{center}
\end{table*}

\subsection{Model Architecture}

We adopt a dual-encoding approach similar to CLIP for processing image and text modalities, leveraging contrastive learning to align these representations. For visual representations, we utilize an encoder-decoder network architecture. Notably, all components of our architecture are trained from scratch without any pretrained initialization. In our vision encoder, we adopt a standard vision transformer (ViT) that divides the input image into $\frac{HW}{P^2}$ patches where $(H,W)$ is the input image resolution and P denotes the patch size. The output tokens corresponding to each input patch are fed into our transformer decoder as shown in Figure \ref{fig:method}. Both text descriptions and captions are processed by a text transformer which utilizes the same architecture employed in CLIP. 

\begin{figure}[t]
    \centering
    \centering
    \includegraphics[width=0.3\textwidth,height=0.3\textwidth]{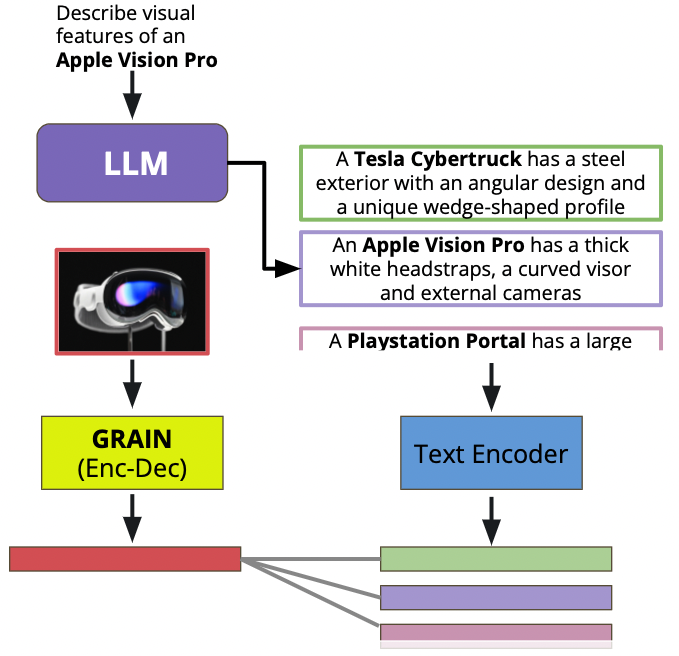}
    \caption{For zero-shot image classification, the image output embedding is compared with text embeddings of classnames enriched with descriptions.}
    \label{fig:inference}
    \vspace{-5mm}
\end{figure}
\vspace{4pt}
\noindent \textbf{Transformer Decoder.} Inspired by DETR~\cite{carion2020endtoend}, we implement a transformer decoder that takes as input a small number of learnable position embeddings called queries and attends to the encoder output. We use two types of queries as input to this model. First we have $n_{q}$ number of queries that we call region queries, whose corresponding outputs are used to predict bounding boxes. Additionally, we use a single image query to learn the overall image context. The transformer model transforms these input queries through self-attention between region and image queries and cross-attention with the encoder output to form output embeddings.  The embeddings corresponding to the region queries are utilized for bounding box prediction and serve as semantic representations for local regions, while the embedding corresponding to the image query captures the overall image representation needed for contrastive learning alongside captions. This image query output is passed through a projection layer before contrastive alignment with the text captions. The bounding box prediction module is exclusively used during training to learn region-aware image features and is inactive during evaluation.%

\vspace{4pt}
\noindent \textbf{Bounding-Box Prediction.} The region output embeddings are fed into a multi-layer perceptron for bounding box prediction. The input size of this MLP is equal to the embedding dimension $d$ and the output size is set to 4, corresponding to the four bounding box coordinates. These MLP weights are shared across all queries.

\vspace{4pt}
\looseness=-1 \noindent \textbf{Semantic Representations.} Each region output embedding is additionally passed through a projection layer to map it into the shared semantic space. The resulting semantic representations are utilized for contrastive alignment with text descriptions. This region-description alignment procedure is illustrated in Figure \ref{fig:alignment}.

\subsection{Training Objectives}
Our approach simultaneously optimizes for three objectives: localizing salient regions within the image, contrastively aligning text descriptions to these salient image region representations, and globally aligning images with captions. 

\begin{table*}[t]
    \centering
    \caption{\small{ Results (Recall@$k$) on zero-shot image-to-text and text-to-image retrieval tasks on MS-COCO and Flickr30k. 
}}
    \small
    \resizebox{1.0\textwidth}{!}{
    \begin{tabular}{lc|cccccc|cccccc}
  
    \toprule[1.2pt]
    \multirow{3}{*}{\bf Data} & \multirow{3}{*}{\bf Model} & \multicolumn{6}{c|}{ MS-COCO}                        & \multicolumn{6}{c}{ Flickr30k}    \\                     
    &                   & \multicolumn{3}{c}{ \bf Image-to-Text} & \multicolumn{3}{c|}{\bf Text-to-Image} & \multicolumn{3}{c}{\bf Image-to-Text} & \multicolumn{3}{c}{\bf Text-to-Image} \\
    &                   &  R@1     &   R@5    &  R@10    &   R@1    &     R@5   &  R@10    &  R@1     &   R@5    &  R@10    &   R@1    &     R@5   &  R@10       \\

    \midrule 
    \small 
    
    \multirow{2}{*}{CC3M} & \hspace{1.2em} \small CLIP & 15.79	& 38.26	& 50.70	& 13.58	& 33.76	& 46.04 & 27.00	& 53.80	& 66.30	& 21.78	& 44.26	& 55.10  \\
    & \hspace{1.2em} \small GRAIN & \bf 38.26	& \bf 65.96 & \bf	77.03 & \bf	28.81 &	\bf 55.86	& \bf 69.00 & \bf 59.90	& \bf 81.80	& \bf 88.40	& \bf 42.82 & 	\bf 68.21 & \bf 76.54 \\
    \multicolumn{2}{c|}{\hspace{5em}$\Delta$} & 
\bf{\textcolor{green}{+22.47}} & \bf{\textcolor{green}{+27.70}} & \bf{\textcolor{green}{+26.33}} & \bf{\textcolor{green}{+15.23}} & \bf{\textcolor{green}{+22.10}} & \bf{\textcolor{green}{+22.96}} & \bf{\textcolor{green}{+32.90}} & \bf{\textcolor{green}{+28.00}} & \bf{\textcolor{green}{+22.10}} & \bf{\textcolor{green}{+21.04}} & \bf{\textcolor{green}{+23.95}} & \bf{\textcolor{green}{+21.44}}
\\

    \midrule 
    \multirow{2}{*}{CC12M} & \hspace{1.2em}\small CLIP & 41.32	& 69.40	& 80.04 &	30.02	& 57.32 &69.65 & 59.60	& 84.70 &	89.90 &	43.63	& 68.75 &	76.77 \\
    &  \hspace{1.2em} \small GRAIN & \bf 58.30	& \bf 83.07	& \bf 89.67	& \bf 42.66	& \bf 70.77 & \bf 80.83 & \bf 78.00	& \bf 94.60	& \bf 97.80	& \bf 59.36	& \bf 80.01	& \bf 85.59 \\
     \multicolumn{2}{c|}{\hspace{5em}$\Delta$} & \bf{\textcolor{green}{+16.98}} & \bf{\textcolor{green}{+13.67}} & \bf{\textcolor{green}{+9.63}} & \bf{\textcolor{green}{+12.64}} & \bf{\textcolor{green}{+13.45}} & \bf{\textcolor{green}{+11.18}} & \bf{\textcolor{green}{+18.40}} & \bf{\textcolor{green}{+9.90}} & \bf{\textcolor{green}{+7.90}} & \bf{\textcolor{green}{+15.73}} & \bf{\textcolor{green}{+11.26}} & \bf{\textcolor{green}{+8.82}} \\

    \bottomrule[1.2pt]
    \end{tabular}
    }
    \label{tab:retrieval}
\end{table*}
\vspace{4pt}
\noindent \textbf{Image-Caption Alignment ($\mathcal{L}_{ic}$).} We adopt the symmetric cross entropy loss from CLIP to maximize the similarity between correct image-caption pairings while contrasting against incorrect pairings within the batch. As with CLIP, we use the [EOS] token from the last layer of the text transformer and the output embedding corresponding to the image query as feature representations for $\mathcal{L}_{ic}$. \

\vspace{4pt}
\noindent \textbf{Bounding Box Loss ($\mathcal{L}_{box}$).} Our model predicts $n_q$ bounding boxes per image corresponding to the region queries. $n_q$ is set to be greater than or equal to the maximum number of objects per image in the training set. Given the variable number of objects per image, we employ the Hungarian Matching algorithm to establish a bipartite matching between predicted and ground truth boxes. For the matched boxes, we implement the bounding box loss derived from DETR, which combines the scale-invariant IOU loss and the L1 loss between the bounding box coordinates. Overall, the bounding box $\mathcal{L}_{box}(b_i,\hat{b}_{\sigma(i)})$ is defined as $\mathcal{L}_{iou}(b_{i},\hat{b}_{\sigma(i)}) + \lVert b_{i} - \hat{b}_{\sigma(i)} \rVert_{1}$. 

\vspace{4pt}
\noindent \textbf{Region-Description Alignment ($\mathcal{L}_{rd}$).} We use an InfoNCE loss \cite{oord2019representation} to learn alignment between output region embeddings and descriptions. Here, the descriptions corresponding to ground truth bounding boxes serve as supervision. We leverage the matched indices between predicted outputs and ground truth boxes obtained via the Hungarian Matching algorithm in the last step to determine ground-truth descriptions for each predicted region output embedding. These matched ground truths are considered positive pairings, while all other pairings within the batch are treated as negatives for InfoNCE. Optimizing for this loss enables our model to learn fine-grained associations between rich textual descriptions and salient image regions that contain discriminative visual features. 
Overall, the final objective function is an equally weighted combination of three components.
\begin{align}
    \mathcal{L}_{total} = \mathcal{L}_{ic} + \mathcal{L}_{box} + \mathcal{L}_{rd}
\end{align}
\vspace{-3mm}
\subsection{Inference} At inference time, our model behaves similar to CLIP, conducting zero-shot classification/retrieval by computing image-text similarities. The image output embedding from our decoder serves as the feature representation for the image. Through self and cross-attention mechanisms, this feature is informed about the fine-grained regions that are characteristic of the given image. The localization modules are inactive during inference; however, they can be used to provide valuable insights for interpreting the model's predictions. For zero-shot image classification (Tables \ref{table:zeroshot-main}, \ref{tab:products}), we enhance class names by appending their descriptions, as illustrated in Figure \ref{fig:inference}. These descriptions are sourced from a LLM similar to \cite{pratt2023does,menon2023visual}. Leveraging the rich image-text correspondences learned during training, our model effectively uses these descriptions to recognize fine-grained and novel categories. %

\section{Experiments}
\begin{table*}[t]
\begin{center}
\caption{\small{We report top-1 accuracy (\%) for zero-shot attribute-based classification. This is a challenging task as indicated by the results.
}}
\vspace*{1.5mm}
\label{table:zeroshot-attr}
\resizebox{1.0\textwidth}{!}{
\begin{tabular}{c@{\hspace{1.7em}}c@{\hspace{0.5em}}|@{\hspace{0.5em}}c@{\hspace{0.5em}}c@{\hspace{0.5em}}c@{\hspace{0.5em}}c@{\hspace{0.5em}}c@{\hspace{0.5em}}c@{\hspace{0.5em}}c@{\hspace{0.5em}}c@{\hspace{0.5em}}c@{\hspace{0.5em}}c@{\hspace{0.5em}}c@{\hspace{0.5em}}c@{\hspace{0.5em}}|@{\hspace{0.5em}}c@{\hspace{0.5em}}}

\toprule[1.2pt]
\bf Data&\bf Model&
\rotatebox[origin=lb]{90}{\smash{\small CIFAR-10}} & \rotatebox[origin=lb]{90}{\smash{\small CIFAR-100}} &
\rotatebox[origin=lb]{90}{\smash{\small SUN397}} & \rotatebox[origin=lb]{90}{\smash{\small Cars}} & \rotatebox[origin=lb]{90}{\smash{\small DTD}} & \rotatebox[origin=lb]{90}{\smash{\small Pets}} & \rotatebox[origin=lb]{90}{\smash{\small Caltech-101}} &
\rotatebox[origin=lb]{90}{\smash{\small Flowers}} & \rotatebox[origin=lb]{90}{\smash{\small CUB}} &  \rotatebox[origin=lb]{90}{\smash{\small Places365}} & \rotatebox[origin=lb]{90}{\smash{\small Food101}}  & \rotatebox[origin=lb]{90}{\smash{\small \bf Average}} & \rotatebox[origin=lb]{90}{\smash{\small \bf ImageNet}} \\
\midrule
\multirow{3}{1.5em}{\rotatebox[origin=c]{0}{\small CC3M}}  & \small CLIP &  24.20 & 7.30 &  13.65 & 0.75 & 6.86 & 3.43 & 24.68 & 1.90 & \bf 1.79 &  8.93 & 5.04 & 8.97 & 4.53 \\
                                                    & \small GRAIN (Ours) & \bf  46.06 & \bf 18.20 &  \bf 20.02 & \bf 0.95 & \bf 14.57 & \bf 4.87 & \bf 45.82 & \bf 2.34 &  1.72 & \bf 13.06 & \bf 7.63 & \bf 15.93 & \bf 7.87 \\
                                                    & $\Delta$ & \bf{\textcolor{green}{+21.86}} & \bf{\textcolor{green}{+10.90}} & \bf{\textcolor{green}{+6.37}} & \bf{\textcolor{green}{+0.20}} & \bf{\textcolor{green}{+7.71}} & \bf{\textcolor{green}{+1.44}} & \bf{\textcolor{green}{+21.14}} & \bf{\textcolor{green}{+0.44}} & \bf{\textcolor{red}{-0.07}} & \bf{\textcolor{green}{+4.13}} & \bf{\textcolor{green}{+2.59}} & \bf{\textcolor{green}{+6.96}} & \bf{\textcolor{green}{+3.34}} \\
\midrule
\multirow{3}{1.8em}{\rotatebox[origin=c]{0}{\small CC12M}}  & \small CLIP &  43.71 &  16.05 & 23.06 & 1.67 & 11.33 & 7.02 & 40.61 & \bf 4.08 & 2.29 & 14.78 & 12.74 & 16.12 & 9.41 \\
                                                    & \small GRAIN (Ours) & \bf 67.39 &  \bf 26.29 & \bf 32.46 & \bf 4.21 & \bf 17.61 & \bf 12.38 & \bf 59.09 &  3.66 & \bf 2.72 &  \bf 20.39  & \bf 18.29 & \bf 24.04 & \bf 14.53 \\
                                                    & $\Delta$ & \bf{\textcolor{green}{+23.68}} & \bf{\textcolor{green}{+10.24}} & \bf{\textcolor{green}{+9.40}} & \bf{\textcolor{green}{+2.54}} & \bf{\textcolor{green}{+6.28}} & \bf{\textcolor{green}{+5.36}} & \bf{\textcolor{green}{+18.48}} & \bf{\textcolor{red}{-0.42}} & \bf{\textcolor{green}{+0.43}} & \bf{\textcolor{green}{+5.61}} & \bf{\textcolor{green}{+5.55}} & \bf{\textcolor{green}{+7.92}} & \bf{\textcolor{green}{+5.12}} \\

\bottomrule[1.2pt]
\end{tabular}}
\end{center}
\end{table*}

The goal of our method is to learn fine-grained vision-language representations that can aid zero-shot visual recognition. By recognizing and addressing the alignment discrepancy between CLIP’s representations of image regions and the rich textual context, our method learns visual representations that are aware of the salient regions in the image and their associations with corresponding textual descriptions. Although the focus of our method is on visual recognition, we observe that our learned representations are of high quality through experiments on cross-modal retrieval benchmarks. We compare against CLIP as our primary baseline, along with recent works like Menon \& Vondrick \cite{menon2023visual} and CuPL \cite{pratt2023does}, that also leverage complementary information from foundation models to improve upon CLIP. We train all CLIP-based baselines from scratch under the same training conditions and evaluate all approaches with a zero-shot evaluation protocol.

\subsection{Experimental Setup}

\looseness=-1 \textbf{Model Architectures.} For all models, we employ the ViT-B/16 \cite{dosovitskiy2020image} architecture for the vision encoders and the Transformer base model \cite{vaswani2017attention} for text encoders as described in CLIP \cite{radford2021learning}. We include results for additional ViT sizes in the Appendix. Our approach, GRAIN, additionally utilizes a query-decoder with 6 transformer decoder layers. We set the number of queries $n_q$ to 10. The outputs from the decoder are processed by projection layers to obtain features in the semantic space, and a 2-layered MLP for predicting bounding boxes. In addition to these comparisons, we evaluate our approach against the substantially larger LLaVA v1.6 model, which includes a ViT-L/14 paired with Vicuna-13 LLM. For this model, we utilize a pretrained checkpoint from huggingface \cite{wolf2020huggingfaces}.

\vspace{4pt}
\noindent \textbf{Pretraining Setup.} 
All models are pre-trained on two distinct image-text datasets that vary in scale: Conceptual Captions 3M (CC3M) and Conceptual Captions 12M (CC12M)~\cite{sharma2018conceptual}. %
Training for all models is conducted using the AdamW optimizer~\cite{loshchilov2017decoupled} across 35 epochs, using a cosine learning rate schedule and weight decay regularization. We use a batch size of 1024 for CC3M experiments and 2048 for CC12M. Training GRAIN for CC3M on a 8 NVIDIA H100 DGX machine takes about 16 hours and for CC12M experiments on 2 $\times$ 8 H100 machines takes 36 hours. While training GRAIN, we randomly choose between the original caption and the MLLM-generated caption as the text supervision. 

\noindent \textbf{Baselines.} To ensure fair evaluation, all baselines were trained under  conditions similar to GRAIN. The introduction of the decoder architecture in our model results in a 22\% increase in parameter count compared to CLIP. For a more fair comparison we report numbers for CLIP by leveraging the same architectures as GRAIN but with localization modules turned off. This baseline is reported as CLIP* throughout the paper. Additionally, we report the performance of the LLaVA v1.6 model to benchmark our model's performance against a state-of-the-art MLLM. Open-ended MLLMs like LLaVA are known to struggle with fine-grained visual recognition \cite{zhang2024visuallygroundedlanguagemodelsbad}. Hence, we propose a new inference strategy to evaluate LLaVA on classification tasks, providing a stronger baseline. Specifically, we first prompt LLaVA to predict a category for an image. Due to its open-ended nature, we cannot directly determine if the generated answer matches the ground truth. To address this, we use a pretrained CLIP text encoder to map LLaVA's generated answer to the closest category within the dataset's vocabulary. This mapped category is then used as the prediction to compute the top-1 accuracy. We refer to this baseline as \textbf{LLaVA + CLIP} in Table \ref{table:zeroshot-main}, representing a stronger and improved baseline over LLaVA alone. Despite possesing orders of magnitude more parameters and being trained on billion-scale datasets, our method manages to surpass LLaVA's performance%
, which shows that our improvements emerge from careful modeling decisions, rather than a simple increase in data volume or model size. 

\subsection{Zero-shot image classification}

We perform zero-shot classification and evaluate all models on Imagenet and 11 additional datasets encompassing common and fine-grained sets. We measure the top-1 accuracy and report results in Table~\ref{table:zeroshot-main}. Our approach, GRAIN, consistently outperforms the current state-of-the-art across all settings and datasets. Specifically, GRAIN improves the zero-shot performance by as much as \textbf{9\%} in absolute accuracy on Imagenet and achieves similar improvements averaged across all other datasets. Notably, our method surpasses existing benchmarks by significant margins across both fine and coarse-grained datasets, with our most substantial improvement reaching up to \textbf{22\%} absolute accuracy on the Caltech-101~\cite{fei2004learning} dataset within the CC3M training setting.
\begin{table*}[t]
\begin{center}
\caption{\small{Ablation studies on our CC3M trained model reporting top-1 accuracy (\%)%
}}
\vspace*{1.5mm}
\label{table:zeroshot-ablations}
\resizebox{1.0\textwidth}{!}{
\begin{tabular}{c@{\hspace{0.5em}}|@{\hspace{0.5em}}c@{\hspace{0.5em}}c@{\hspace{0.5em}}c@{\hspace{0.5em}}c@{\hspace{0.5em}}c@{\hspace{0.5em}}c@{\hspace{0.5em}}c@{\hspace{0.5em}}c@{\hspace{0.5em}}c@{\hspace{0.5em}}c@{\hspace{0.5em}}c@{\hspace{0.5em}}c@{\hspace{0.5em}}|@{\hspace{0.5em}}c@{\hspace{0.5em}}}

\toprule[1.2pt]
 \bf Setting&
\rotatebox[origin=lb]{90}{\smash{\small CIFAR-10}} & \rotatebox[origin=lb]{90}{\smash{\small CIFAR-100}} &
\rotatebox[origin=lb]{90}{\smash{\small SUN397}} & \rotatebox[origin=lb]{90}{\smash{\small Cars}} & \rotatebox[origin=lb]{90}{\smash{\small DTD}} & \rotatebox[origin=lb]{90}{\smash{\small Pets}} & \rotatebox[origin=lb]{90}{\smash{\small Caltech-101}} &
\rotatebox[origin=lb]{90}{\smash{\small Flowers}} & \rotatebox[origin=lb]{90}{\smash{\small CUB}} &  \rotatebox[origin=lb]{90}{\smash{\small Places365}} & \rotatebox[origin=lb]{90}{\smash{\small Food101}}  & \rotatebox[origin=lb]{90}{\smash{\small  \bf Average}} & \rotatebox[origin=lb]{90}{\smash{\small  \bf ImageNet}} \\
\midrule
\multirow{1}{13em}{\rotatebox[origin=c]{0}\tiny \hspace{0.5em} GRAIN}&  \bf 65.86 & \bf 35.20 &  \bf 38.07 & \bf 1.34 & \bf 17.24 & \bf 14.15 & \bf 65.20 & \bf 13.24 & \bf 5.47 & \bf 24.96 & \bf 16.18 & \bf 27.00 & \bf 23.34 \\
                                                  \multirow{1}{13em} {\rotatebox[origin=c]{0}\tiny \hspace{0.5em} -- Region-description loss} &  58.21 & 27.07 & 35.28  &  1.01 & 14.20 & 9.18 & 58.86 & 9.13 & 3.52 &  22.31  & 13.05 & 22.89 & 18.73 \\
                                                    \multirow{1}{13em}{\rotatebox[origin=c]{0} \tiny \hspace{1.0em} -- Box loss} &  57.06 & 26.17 & 34.38  &  0.93 & 14.67 & 8.87 & 56.91 & 8.31 & 3.20 &  21.35  & 13.12 & 22.27 & 17.54 \\
                                                  \multirow{1}{13em} {\rotatebox[origin=c]{0}\tiny \hspace{1.5em} -- MLLM-caption} & 47.24 & 19.92 & 28.51 & 0.70 & 8.78 & 7.04 & 43.95 & 8.20 & 2.99 &  20.06  & 9.01 & 17.85 & 14.56 \\
                                                     \multirow{1}{13em}{\rotatebox[origin=c]{0}\tiny \hspace{2.0em} -- Menon\&Vondrick \cite{menon2023visual}} & 46.99 & 18.49 & 29.76 & 0.52 & 8.40 & 6.62 & 42.56 & 8.29 & 3.36 &  18.70  & 10.01 & 17.62 & 14.04 \\

\bottomrule[1.2pt]
\end{tabular}}
\vspace{-0.5cm}
\end{center}
\end{table*}

\subsection{Cross-modal retrieval}
We evaluate the pre-trained models on the task of cross-modal retrieval under the zero-shot setting. Specifically, we focus on the Image-to-Text (I2T) and Text-to-Image (T2I) retrieval tasks using the MSCOCO and Flickr30k datasets in Table \ref{tab:retrieval}. Our evaluations are conducted on the standard test sets for both datasets, and we report performance metrics in terms of Recall@k for k values of 1, 5, and 10. Compared to CLIP, our method achieves superior performance with performance gains of up to \textbf{~33\%}. On average, we observe improvements of \textbf{23.8\%} with CC3M trained models and \textbf{12.46\%} with models trained on CC12M.
\subsection{Zero-shot attribute-based classification}
To measure image-description alignment, we design an experiment to classify images by leveraging only descriptions/attributes. This is a challenging task, as image classification is being performed devoid of class names. 
Toward this end, we first prompted GPT-3 using class names from the downstream dataset's vocabulary to obtain descriptions. Next, instead of the traditional approach of encoding class names and computing similarities with images, we encoded the description corresponding to the class name (omitting the class name itself) to obtain the text representation and computed similarities with images. The class corresponding to the text representation that scored the maximum similarity with the test image is considered the prediction for that image. We compute top-1 accuracy as usual and reported for all datasets in Table \ref{table:zeroshot-attr}. From Table \ref{table:zeroshot-attr}, we observe that our model is able to achieve strong improvements over CLIP, demonstrating closer image-description alignment. On average, we achieve an improvement of \textbf{6-7\%} over CLIP, showcasing better alignment.

\subsection{Recognizing Novel Examples}
It is desirable for open-vocabulary models to generalize to novel, unseen examples at test-time without requiring re-training. Zero-shot learning methods often utilize auxiliary information, such as attributes, for classifying unknown entities. Hence, our approach aims to recognize these concepts by leveraging LLM-generated descriptions.
 \begingroup
\setlength{\columnsep}{7pt}
\begin{wraptable}{r}{0.27\textwidth}
\vspace{-10pt}
\centering
\resizebox{0.27\textwidth}{!}{%
\begin{tabular}{@{}c|cc@{}}
\hline
\textbf{Accuracy (\%)} & \textbf{Products-2023} \\
\hline
CLIP & 33.65 \\
\hline
 \textcolor{gray}{LLaVA} &  \textcolor{gray}{42.08} \\
\hline
GRAIN & \textbf{45.24} \\
\hline
\end{tabular}
}
\vspace{-8pt}
\caption{}
\label{tab:products}
\vspace{-16pt}
\end{wraptable}
 In this experiment, we aim to test our model's ability in recognizing novel entities that were absent from the training distribution. Toward this end, we collect 1500 images of products launched after 2023, manually filter these images for quality control and label them into 27 novel categories to form a new benchmark dataset. We call this the \textbf{Products-2023} dataset. These concepts are absent from our model's training distribution making them novel. %
 We provide additional details on this dataset in Appendix. We evaluate our model along with CLIP and LLaVA on this dataset in Table \ref{tab:products} which demonstrates superior results achieved by our approach against CLIP and even against the much larger LLaVA model confirming the efficacy of our approach in recognizing novel samples.

\subsection{Ablations}

To assess the importance of the different components in GRAIN, we conduct four ablation experiments. We restrict to models trained on CC3M due to computational constraints. The outcomes of these ablations are reported as top-1 accuracy in Table \ref{table:zeroshot-ablations}. 

\vspace{4pt}
\noindent \textbf{Ablating the region-description alignment loss.} This component is pivotal to our framework as removing it causes a significant accuracy decline of ~5\% on all datasets on average. This considerable decrease underscores the vital role of this loss in establishing fine-grained correspondences between salient image regions and their descriptions.  

\vspace{4pt}
\noindent \textbf{Ablating the localization loss.} Further removing the bounding box prediction losses from our training regime leads to a modest performance drop. This loss is instrumental in identifying and predicting salient regions within the image, and, in conjunction with the alignment loss is crucial to developing fine-grained visual understanding. 

\vspace{4pt}
\noindent \textbf{Ablating the role of MLLM-caption during training.} We employ captions generated by LLaVA as a form of text-level data augmentation during training, alternating between these and the original image captions. The MLLM-generated caption provides a high-level visual summary of the image, proving to be significant for training, as indicated by a 3\% decrease in performance upon its removal.

\vspace{4pt}
\noindent \textbf{Ablating the role of test-time descriptions.}  In line with the approach of Menon \& Vondrick~\cite{menon2023visual}, we utilize descriptions generated by GPT-3 to enrich class names during zero-shot classification. Excluding these augmented descriptions results in a minor performance reduction, suggesting that while beneficial, our model's performance is not reliant on these test-time descriptions.

\section{Conclusion}
\looseness=-1 In this paper, we propose a new pre-training method for contrastive vision-language models. Specifically, we hypothesize that many of the current limitations of CLIP stem from its image-level contrastive pre-training, which neglects fine-grained alignment. As a result, we propose to leverage Multi-Modal Large Language Models (LLaVA) and Open-Vocabulary Object Detectors (OWLv2) to automatically generate weak supervision to drive a more fine-grained pre-training process. %
We demonstrate superior performance across 11 different classification datasets, including ones containing fine-grained and novel examples, as well as additional tasks such as cross-modal retrieval. Our results show significant improvement over the state-of-art, including by up to 9\% in absolute top-1 accuracy for zero-shot classification and 25\% on retrieval. Our method can even outperform LLaVA, which is over 13B parameters (compared to our $\sim$170M) and was trained on billions of data-points.

{
    \small
    \bibliographystyle{ieeenat_fullname}
    \bibliography{main}

\begin{thebibliography}{52}
\providecommand{\natexlab}[1]{#1}
\providecommand{\url}[1]{\texttt{#1}}
\expandafter\ifx\csname urlstyle\endcsname\relax
  \providecommand{\doi}[1]{doi: #1}\else
  \providecommand{\doi}{doi: \begingroup \urlstyle{rm}\Url}\fi

\bibitem[Bica et~al.(2024)Bica, Ilić, Bauer, Erdogan, Bošnjak, Kaplanis, Gritsenko, Minderer, Blundell, Pascanu, and Mitrović]{bica2024improving}
Ioana Bica, Anastasija Ilić, Matthias Bauer, Goker Erdogan, Matko Bošnjak, Christos Kaplanis, Alexey~A. Gritsenko, Matthias Minderer, Charles Blundell, Razvan Pascanu, and Jovana Mitrović.
\newblock Improving fine-grained understanding in image-text pre-training, 2024.

\bibitem[Brown et~al.(2020)Brown, Mann, Ryder, Subbiah, Kaplan, Dhariwal, Neelakantan, Shyam, Sastry, Askell, Agarwal, Herbert-Voss, Krueger, Henighan, Child, Ramesh, Ziegler, Wu, Winter, Hesse, Chen, Sigler, Litwin, Gray, Chess, Clark, Berner, McCandlish, Radford, Sutskever, and Amodei]{brown2020language}
Tom~B. Brown, Benjamin Mann, Nick Ryder, Melanie Subbiah, Jared Kaplan, Prafulla Dhariwal, Arvind Neelakantan, Pranav Shyam, Girish Sastry, Amanda Askell, Sandhini Agarwal, Ariel Herbert-Voss, Gretchen Krueger, Tom Henighan, Rewon Child, Aditya Ramesh, Daniel~M. Ziegler, Jeffrey Wu, Clemens Winter, Christopher Hesse, Mark Chen, Eric Sigler, Mateusz Litwin, Scott Gray, Benjamin Chess, Jack Clark, Christopher Berner, Sam McCandlish, Alec Radford, Ilya Sutskever, and Dario Amodei.
\newblock Language models are few-shot learners, 2020.

\bibitem[Burns et~al.(2023)Burns, Witzel, Hamid, Yu, Finn, and Hausman]{burns2023makes}
Kaylee Burns, Zach Witzel, Jubayer~Ibn Hamid, Tianhe Yu, Chelsea Finn, and Karol Hausman.
\newblock What makes pre-trained visual representations successful for robust manipulation?, 2023.

\bibitem[Carion et~al.(2020)Carion, Massa, Synnaeve, Usunier, Kirillov, and Zagoruyko]{carion2020endtoend}
Nicolas Carion, Francisco Massa, Gabriel Synnaeve, Nicolas Usunier, Alexander Kirillov, and Sergey Zagoruyko.
\newblock End-to-end object detection with transformers, 2020.

\bibitem[Chen et~al.(2023)Chen, Shvetsova, Rouditchenko, Kondermann, Thomas, Chang, Feris, Glass, and Kuehne]{chen2023what}
Brian Chen, Nina Shvetsova, Andrew Rouditchenko, Daniel Kondermann, Samuel Thomas, Shih-Fu Chang, Rogerio Feris, James Glass, and Hilde Kuehne.
\newblock What, when, and where? -- self-supervised spatio-temporal grounding in untrimmed multi-action videos from narrated instructions, 2023.

\bibitem[Chen et~al.(2020)Chen, Li, Yu, Kholy, Ahmed, Gan, Cheng, and Liu]{chen2020uniter}
Yen-Chun Chen, Linjie Li, Licheng Yu, Ahmed~El Kholy, Faisal Ahmed, Zhe Gan, Yu Cheng, and Jingjing Liu.
\newblock Uniter: Universal image-text representation learning, 2020.

\bibitem[Chiang et~al.(2023)Chiang, Li, Lin, Sheng, Wu, Zhang, Zheng, Zhuang, Zhuang, Gonzalez, Stoica, and Xing]{vicuna2023}
Wei-Lin Chiang, Zhuohan Li, Zi Lin, Ying Sheng, Zhanghao Wu, Hao Zhang, Lianmin Zheng, Siyuan Zhuang, Yonghao Zhuang, Joseph~E. Gonzalez, Ion Stoica, and Eric~P. Xing.
\newblock Vicuna: An open-source chatbot impressing gpt-4 with 90\%* chatgpt quality, 2023.

\bibitem[Conti et~al.(2023)Conti, Fini, Mancini, Rota, Wang, and Ricci]{conti2023vocabularyfree}
Alessandro Conti, Enrico Fini, Massimiliano Mancini, Paolo Rota, Yiming Wang, and Elisa Ricci.
\newblock Vocabulary-free image classification, 2023.

\bibitem[Dong et~al.(2023)Dong, Han, Peng, Qi, Ge, Yang, Zhao, Sun, Zhou, Wei, et~al.]{dong2023dreamllm}
Runpei Dong, Chunrui Han, Yuang Peng, Zekun Qi, Zheng Ge, Jinrong Yang, Liang Zhao, Jianjian Sun, Hongyu Zhou, Haoran Wei, et~al.
\newblock Dreamllm: Synergistic multimodal comprehension and creation.
\newblock \emph{arXiv preprint arXiv:2309.11499}, 2023.

\bibitem[Dosovitskiy et~al.(2020)Dosovitskiy, Beyer, Kolesnikov, Weissenborn, Zhai, Unterthiner, Dehghani, Minderer, Heigold, Gelly, et~al.]{dosovitskiy2020image}
Alexey Dosovitskiy, Lucas Beyer, Alexander Kolesnikov, Dirk Weissenborn, Xiaohua Zhai, Thomas Unterthiner, Mostafa Dehghani, Matthias Minderer, Georg Heigold, Sylvain Gelly, et~al.
\newblock An image is worth 16x16 words: Transformers for image recognition at scale.
\newblock \emph{arXiv preprint arXiv:2010.11929}, 2020.

\bibitem[Fan et~al.(2023)Fan, Krishnan, Isola, Katabi, and Tian]{fan2023improving}
Lijie Fan, Dilip Krishnan, Phillip Isola, Dina Katabi, and Yonglong Tian.
\newblock Improving clip training with language rewrites.
\newblock In \emph{NeurIPS}, 2023.

\bibitem[Fei-Fei et~al.(2004)Fei-Fei, Fergus, and Perona]{fei2004learning}
Li Fei-Fei, Rob Fergus, and Pietro Perona.
\newblock Learning generative visual models from few training examples: An incremental bayesian approach tested on 101 object categories.
\newblock In \emph{2004 conference on computer vision and pattern recognition workshop}, pages 178--178. IEEE, 2004.

\bibitem[Hu et~al.(2023)Hu, Luan, Chen, Khandelwal, Joshi, Lee, Toutanova, and Chang]{hu2023open}
Hexiang Hu, Yi Luan, Yang Chen, Urvashi Khandelwal, Mandar Joshi, Kenton Lee, Kristina Toutanova, and Ming-Wei Chang.
\newblock Open-domain visual entity recognition: Towards recognizing millions of wikipedia entities.
\newblock \emph{International Conference on Computer Vision}, 2023.

\bibitem[Huang et~al.(2021)Huang, Shen, Lungren, and Yeung]{9710099}
Shih-Cheng Huang, Liyue Shen, Matthew~P. Lungren, and Serena Yeung.
\newblock Gloria: A multimodal global-local representation learning framework for label-efficient medical image recognition.
\newblock In \emph{2021 IEEE/CVF International Conference on Computer Vision (ICCV)}, pages 3922--3931, 2021.

\bibitem[Jia et~al.(2021)Jia, Yang, Xia, Chen, Parekh, Pham, Le, Sung, Li, and Duerig]{jia2021scaling}
Chao Jia, Yinfei Yang, Ye Xia, Yi-Ting Chen, Zarana Parekh, Hieu Pham, Quoc~V. Le, Yunhsuan Sung, Zhen Li, and Tom Duerig.
\newblock Scaling up visual and vision-language representation learning with noisy text supervision, 2021.

\bibitem[Kamath et~al.(2021)Kamath, Singh, LeCun, Synnaeve, Misra, and Carion]{kamath2021mdetr}
Aishwarya Kamath, Mannat Singh, Yann LeCun, Gabriel Synnaeve, Ishan Misra, and Nicolas Carion.
\newblock Mdetr -- modulated detection for end-to-end multi-modal understanding, 2021.

\bibitem[Lai et~al.(2024)Lai, Zhang, Zhang, Wu, Bai, Timofeev, Du, Gan, Shan, Chuah, Yang, and Cao]{lai2024veclipimprovingcliptraining}
Zhengfeng Lai, Haotian Zhang, Bowen Zhang, Wentao Wu, Haoping Bai, Aleksei Timofeev, Xianzhi Du, Zhe Gan, Jiulong Shan, Chen-Nee Chuah, Yinfei Yang, and Meng Cao.
\newblock Veclip: Improving clip training via visual-enriched captions, 2024.

\bibitem[Lampert et~al.(2009)Lampert, Nickisch, and Harmeling]{5206594}
Christoph~H. Lampert, Hannes Nickisch, and Stefan Harmeling.
\newblock Learning to detect unseen object classes by between-class attribute transfer.
\newblock In \emph{2009 IEEE Conference on Computer Vision and Pattern Recognition}, pages 951--958, 2009.

\bibitem[Lei et~al.(2021)Lei, Berg, and Bansal]{lei2021qvhighlights}
Jie Lei, Tamara~L. Berg, and Mohit Bansal.
\newblock Qvhighlights: Detecting moments and highlights in videos via natural language queries, 2021.

\bibitem[Li et~al.(2020)Li, Yin, Li, Zhang, Hu, Zhang, Wang, Hu, Dong, Wei, Choi, and Gao]{li2020oscar}
Xiujun Li, Xi Yin, Chunyuan Li, Pengchuan Zhang, Xiaowei Hu, Lei Zhang, Lijuan Wang, Houdong Hu, Li Dong, Furu Wei, Yejin Choi, and Jianfeng Gao.
\newblock Oscar: Object-semantics aligned pre-training for vision-language tasks, 2020.

\bibitem[Liu et~al.(2024)Liu, Li, Wu, and Lee]{liu2024visual}
Haotian Liu, Chunyuan Li, Qingyang Wu, and Yong~Jae Lee.
\newblock Visual instruction tuning.
\newblock \emph{Advances in neural information processing systems}, 36, 2024.

\bibitem[Loshchilov and Hutter(2017)]{loshchilov2017decoupled}
Ilya Loshchilov and Frank Hutter.
\newblock Decoupled weight decay regularization.
\newblock \emph{arXiv preprint arXiv:1711.05101}, 2017.

\bibitem[Lu et~al.(2019)Lu, Batra, Parikh, and Lee]{lu2019vilbert}
Jiasen Lu, Dhruv Batra, Devi Parikh, and Stefan Lee.
\newblock Vilbert: Pretraining task-agnostic visiolinguistic representations for vision-and-language tasks, 2019.

\bibitem[Maji et~al.(2013)Maji, Rahtu, Kannala, Blaschko, and Vedaldi]{maji2013finegrained}
Subhransu Maji, Esa Rahtu, Juho Kannala, Matthew Blaschko, and Andrea Vedaldi.
\newblock Fine-grained visual classification of aircraft, 2013.

\bibitem[Menon and Vondrick(2023)]{menon2023visual}
Sachit Menon and Carl Vondrick.
\newblock Visual classification via description from large language models.
\newblock In \emph{The Eleventh International Conference on Learning Representations}, 2023.

\bibitem[Minderer et~al.(2024)Minderer, Gritsenko, and Houlsby]{minderer2024scaling}
Matthias Minderer, Alexey Gritsenko, and Neil Houlsby.
\newblock Scaling open-vocabulary object detection.
\newblock \emph{Advances in Neural Information Processing Systems}, 36, 2024.

\bibitem[Mirza et~al.(2023)Mirza, Karlinsky, Lin, Possegger, Feris, and Bischof]{mirza2023tap}
M.~Jehanzeb Mirza, Leonid Karlinsky, Wei Lin, Horst Possegger, Rogerio Feris, and Horst Bischof.
\newblock Tap: Targeted prompting for task adaptive generation of textual training instances for visual classification, 2023.

\bibitem[Mu et~al.(2021)Mu, Kirillov, Wagner, and Xie]{mu2021slip}
Norman Mu, Alexander Kirillov, David Wagner, and Saining Xie.
\newblock Slip: Self-supervision meets language-image pre-training, 2021.

\bibitem[OpenAI(2023)]{gpt4v}
OpenAI.
\newblock Gpt-4v(ision) system card.
\newblock \emph{OpenAI}, 2023.

\bibitem[Parcalabescu et~al.(2022)Parcalabescu, Cafagna, Muradjan, Frank, Calixto, and Gatt]{parcalabescu-etal-2022-valse}
Letitia Parcalabescu, Michele Cafagna, Lilitta Muradjan, Anette Frank, Iacer Calixto, and Albert Gatt.
\newblock {VALSE}: A task-independent benchmark for vision and language models centered on linguistic phenomena.
\newblock In \emph{Proceedings of the 60th Annual Meeting of the Association for Computational Linguistics (Volume 1: Long Papers)}, pages 8253--8280, Dublin, Ireland, 2022. Association for Computational Linguistics.

\bibitem[Parikh and Grauman(2011)]{6126281}
Devi Parikh and Kristen Grauman.
\newblock Relative attributes.
\newblock In \emph{2011 International Conference on Computer Vision}, pages 503--510, 2011.

\bibitem[Pratt et~al.(2023)Pratt, Covert, Liu, and Farhadi]{pratt2023does}
Sarah Pratt, Ian Covert, Rosanne Liu, and Ali Farhadi.
\newblock What does a platypus look like? generating customized prompts for zero-shot image classification.
\newblock In \emph{Proceedings of the IEEE/CVF International Conference on Computer Vision}, pages 15691--15701, 2023.

\bibitem[Radford et~al.()Radford, Kim, Hallacy, Ramesh, Goh, Agarwal, Sastry, Askell, Mishkin, Clark, et~al.]{radfordlearning}
Alec Radford, Jong~Wook Kim, Chris Hallacy, Aditya Ramesh, Gabriel Goh, Sandhini Agarwal, Girish Sastry, Amanda Askell, Pamela Mishkin, Jack Clark, et~al.
\newblock Learning transferable visual models from natural language supervision.

\bibitem[Radford et~al.(2021)Radford, Kim, Hallacy, Ramesh, Goh, Agarwal, Sastry, Askell, Mishkin, Clark, et~al.]{radford2021learning}
Alec Radford, Jong~Wook Kim, Chris Hallacy, Aditya Ramesh, Gabriel Goh, Sandhini Agarwal, Girish Sastry, Amanda Askell, Pamela Mishkin, Jack Clark, et~al.
\newblock Learning transferable visual models from natural language supervision.
\newblock In \emph{International conference on machine learning}, pages 8748--8763. PMLR, 2021.

\bibitem[Ranasinghe et~al.(2023)Ranasinghe, McKinzie, Ravi, Yang, Toshev, and Shlens]{ranasinghe2023perceptual}
Kanchana Ranasinghe, Brandon McKinzie, Sachin Ravi, Yinfei Yang, Alexander Toshev, and Jonathon Shlens.
\newblock Perceptual grouping in contrastive vision-language models, 2023.

\bibitem[Rasheed et~al.(2023)Rasheed, Maaz, Shaji, Shaker, Khan, Cholakkal, Anwer, Xing, Yang, and Khan]{rasheed2023glamm}
Hanoona Rasheed, Muhammad Maaz, Sahal Shaji, Abdelrahman Shaker, Salman Khan, Hisham Cholakkal, Rao~M Anwer, Erix Xing, Ming-Hsuan Yang, and Fahad~S Khan.
\newblock Glamm: Pixel grounding large multimodal model.
\newblock \emph{arXiv preprint arXiv:2311.03356}, 2023.

\bibitem[Ren et~al.(2023)Ren, Su, and Liu]{ren2023chatgptpowered}
Zhiyuan Ren, Yiyang Su, and Xiaoming Liu.
\newblock Chatgpt-powered hierarchical comparisons for image classification, 2023.

\bibitem[Schuhmann et~al.(2022)Schuhmann, Beaumont, Vencu, Gordon, Wightman, Cherti, Coombes, Katta, Mullis, Wortsman, et~al.]{schuhmann2022laion}
Christoph Schuhmann, Romain Beaumont, Richard Vencu, Cade Gordon, Ross Wightman, Mehdi Cherti, Theo Coombes, Aarush Katta, Clayton Mullis, Mitchell Wortsman, et~al.
\newblock Laion-5b: An open large-scale dataset for training next generation image-text models.
\newblock \emph{Advances in Neural Information Processing Systems}, 35:\penalty0 25278--25294, 2022.

\bibitem[Sharma et~al.(2018)Sharma, Ding, Goodman, and Soricut]{sharma2018conceptual}
Piyush Sharma, Nan Ding, Sebastian Goodman, and Radu Soricut.
\newblock Conceptual captions: A cleaned, hypernymed, image alt-text dataset for automatic image captioning.
\newblock In \emph{Proceedings of the 56th Annual Meeting of the Association for Computational Linguistics (Volume 1: Long Papers)}, pages 2556--2565, 2018.

\bibitem[van~den Oord et~al.(2019)van~den Oord, Li, and Vinyals]{oord2019representation}
Aaron van~den Oord, Yazhe Li, and Oriol Vinyals.
\newblock Representation learning with contrastive predictive coding, 2019.

\bibitem[Vaswani et~al.(2017)Vaswani, Shazeer, Parmar, Uszkoreit, Jones, Gomez, Kaiser, and Polosukhin]{vaswani2017attention}
Ashish Vaswani, Noam Shazeer, Niki Parmar, Jakob Uszkoreit, Llion Jones, Aidan~N Gomez, {\L}ukasz Kaiser, and Illia Polosukhin.
\newblock Attention is all you need.
\newblock \emph{Advances in neural information processing systems}, 30, 2017.

\bibitem[Wah et~al.(2011)Wah, Branson, Welinder, Perona, and Belongie]{WahCUB_200_2011}
C. Wah, S. Branson, P. Welinder, P. Perona, and S. Belongie.
\newblock Cub.
\newblock Technical Report CNS-TR-2011-001, California Institute of Technology, 2011.

\bibitem[Wang et~al.(2022)Wang, Ge, Cai, Yan, Lin, Shan, Qie, and Shou]{wang2022objectaware}
Alex~Jinpeng Wang, Yixiao Ge, Guanyu Cai, Rui Yan, Xudong Lin, Ying Shan, Xiaohu Qie, and Mike~Zheng Shou.
\newblock Object-aware video-language pre-training for retrieval, 2022.

\bibitem[Wolf et~al.(2020)Wolf, Debut, Sanh, Chaumond, Delangue, Moi, Cistac, Rault, Louf, Funtowicz, Davison, Shleifer, von Platen, Ma, Jernite, Plu, Xu, Scao, Gugger, Drame, Lhoest, and Rush]{wolf2020huggingfaces}
Thomas Wolf, Lysandre Debut, Victor Sanh, Julien Chaumond, Clement Delangue, Anthony Moi, Pierric Cistac, Tim Rault, Rémi Louf, Morgan Funtowicz, Joe Davison, Sam Shleifer, Patrick von Platen, Clara Ma, Yacine Jernite, Julien Plu, Canwen Xu, Teven~Le Scao, Sylvain Gugger, Mariama Drame, Quentin Lhoest, and Alexander~M. Rush.
\newblock Huggingface's transformers: State-of-the-art natural language processing, 2020.

\bibitem[Xian et~al.(2018)Xian, Lorenz, Schiele, and Akata]{8578679}
Yongqin Xian, Tobias Lorenz, Bernt Schiele, and Zeynep Akata.
\newblock Feature generating networks for zero-shot learning.
\newblock In \emph{2018 IEEE/CVF Conference on Computer Vision and Pattern Recognition}, pages 5542--5551, 2018.

\bibitem[Yao et~al.(2021)Yao, Huang, Hou, Lu, Niu, Xu, Liang, Li, Jiang, and Xu]{yao2021filip}
Lewei Yao, Runhui Huang, Lu Hou, Guansong Lu, Minzhe Niu, Hang Xu, Xiaodan Liang, Zhenguo Li, Xin Jiang, and Chunjing Xu.
\newblock Filip: Fine-grained interactive language-image pre-training, 2021.

\bibitem[Ye et~al.(2022)Ye, Gao, Li, Xu, Feng, Wu, Yu, and Kong]{ye2022zerogen}
Jiacheng Ye, Jiahui Gao, Qintong Li, Hang Xu, Jiangtao Feng, Zhiyong Wu, Tao Yu, and Lingpeng Kong.
\newblock Zerogen: Efficient zero-shot learning via dataset generation.
\newblock 2022.

\bibitem[Yuksekgonul et~al.(2023)Yuksekgonul, Bianchi, Kalluri, Jurafsky, and Zou]{yuksekgonul2023visionlanguage}
Mert Yuksekgonul, Federico Bianchi, Pratyusha Kalluri, Dan Jurafsky, and James Zou.
\newblock When and why vision-language models behave like bags-of-words, and what to do about it?, 2023.

\bibitem[Zhai et~al.(2023)Zhai, Mustafa, Kolesnikov, and Beyer]{zhai2023sigmoid}
Xiaohua Zhai, Basil Mustafa, Alexander Kolesnikov, and Lucas Beyer.
\newblock Sigmoid loss for language image pre-training, 2023.

\bibitem[Zhang et~al.(2023)Zhang, Gupta, and Zisserman]{zhang2023helping}
Chuhan Zhang, Ankush Gupta, and Andrew Zisserman.
\newblock Helping hands: An object-aware ego-centric video recognition model, 2023.

\bibitem[Zhang et~al.(2024)Zhang, Unell, Wang, Ghosh, Su, Schmidt, and Yeung-Levy]{zhang2024visuallygroundedlanguagemodelsbad}
Yuhui Zhang, Alyssa Unell, Xiaohan Wang, Dhruba Ghosh, Yuchang Su, Ludwig Schmidt, and Serena Yeung-Levy.
\newblock Why are visually-grounded language models bad at image classification?, 2024.

\bibitem[Zhu et~al.(2023)Zhu, Chen, Shen, Li, and Elhoseiny]{zhu2023minigpt}
Deyao Zhu, Jun Chen, Xiaoqian Shen, Xiang Li, and Mohamed Elhoseiny.
\newblock Minigpt-4: Enhancing vision-language understanding with advanced large language models.
\newblock \emph{arXiv preprint arXiv:2304.10592}, 2023.

\end{thebibliography}
}

\clearpage
\appendix
\section*{Appendix}
\setcounter{figure}{0}
\setcounter{table}{0}
\renewcommand{\thetable}{\Alph{table}}
\renewcommand{\thefigure}{\Alph{figure}}
\renewcommand\thesection{\Alph{section}}

\section{Additional Details on Related Work}
The scope of our work spans several domains including Vision-Language Representation Learning, Fine-grained Visual Recognition, Visual Grounding and Open-world/Zero-shot learning. Since the related works section in the main paper cannot adequately cover all of these areas, we provide a more comprehensive summary in this supplementary material:
\begin{table*}[t]
\begin{center}
\caption{\small{Zero-shot top-1 accuracy (\%) of different methods using the \textbf{ViT-L/14 backbone}.
}}
\vspace*{1.5mm}
\label{table:zeroshot-sup}
\resizebox{1.0\textwidth}{!}{
\begin{tabular}{c@{\hspace{1.9em}}c@{\hspace{0.5em}}|@{\hspace{0.5em}}c@{\hspace{0.5em}}c@{\hspace{0.5em}}c@{\hspace{0.5em}}c@{\hspace{0.5em}}c@{\hspace{0.5em}}c@{\hspace{0.5em}}c@{\hspace{0.5em}}c@{\hspace{0.5em}}c@{\hspace{0.5em}}c@{\hspace{0.5em}}c@{\hspace{0.5em}}c@{\hspace{0.5em}}|@{\hspace{0.5em}}c@{\hspace{0.5em}}}

\toprule[1.2pt]
\bf Data&\bf Model&
\rotatebox[origin=lb]{90}{\smash{\small CIFAR-10}} & \rotatebox[origin=lb]{90}{\smash{\small CIFAR-100}} &
\rotatebox[origin=lb]{90}{\smash{\small SUN397}} & \rotatebox[origin=lb]{90}{\smash{\small Cars}} & \rotatebox[origin=lb]{90}{\smash{\small DTD}} & \rotatebox[origin=lb]{90}{\smash{\small Pets}} & \rotatebox[origin=lb]{90}{\smash{\small Caltech-101}} &
\rotatebox[origin=lb]{90}{\smash{\small Flowers}} & \rotatebox[origin=lb]{90}{\smash{\small CUB}} &  \rotatebox[origin=lb]{90}{\smash{\small Places365}} & \rotatebox[origin=lb]{90}{\smash{\small Food101}}  & \rotatebox[origin=lb]{90}{\smash{\small \bf Average}} & \rotatebox[origin=lb]{90}{\smash{\small \bf ImageNet}} \\
\midrule
 \multirow{1}{1.8em}{\rotatebox[origin=c]{0}{\small }}  &  \small \textcolor{gray}{LLaVA + CLIP}&  \textcolor{gray}{89.69} &  \textcolor{gray}{57.72} & \textcolor{gray}{55.24} & \textcolor{gray}{15.90} & \textcolor{gray}{35.37} & \textcolor{gray}{47.16} & \textcolor{gray}{75.03} & \textcolor{gray}{24.69} & \textcolor{gray}{6.22} &  \textcolor{gray}{29.43}  & \textcolor{gray}{52.80} & \textcolor{gray}{44.48} & \textcolor{gray}{35.20}\\    
\midrule
\multirow{4}{1.8em}{\rotatebox[origin=c]{0}{\small CC12M}}   & \small CLIP \cite{radford2021learning} &  73.36 &  38.06 & 49.96 & 4.59 & 21.84 & 43.98 & 71.79 & 22.01 & 7.72 & 33.16 & 42.25 & 37.15 & 36.72 \\
& \small Menon\&Vondrick \cite{menon2023visual} &  73.74 & 38.48 &  50.05 & 5.22 & 22.04 & 44.33 & 72.56 & 22.10 & 8.28 &  33.78  & 43.04 & 37.60 & 36.84 \\
                                                    & \small CuPL \cite{pratt2023does} & 73.53 & 38.55 &  50.46 & 5.14 & 21.96 & 43.28 & 73.08 & 23.12 & 8.65 &  32.48  & 42.96 & 37.56 & 37.05 \\
                                                    \cmidrule{2-15}
                                                    & \small GRAIN (Ours) & \bf 81.62 &  \bf 44.98 & \bf 55.82 & \bf 9.12 & \bf 27.66 & \bf 52.98 & \bf 82.05 & \bf 28.18 & \bf 12.73 &  \bf 37.34  & \bf 46.92 & \bf 43.58 & \bf 42.68 \\

\bottomrule[1.2pt]
\end{tabular}}
\end{center}
\end{table*}

\vspace{4pt}
\noindent\textbf{Contrastive Language-Image Pretraining.}
Methods like CLIP \cite{radford2021learning} and ALIGN \cite{jia2021scaling} leverage large internet scraped datasets of image-text pairs to learn a joint representation by contrastively aligning the two modalities. The objective of these methods is to pull together image and text representations that are semantically similar and push apart dissimilar pairs. These works employ a dual encoder approach, separately encoding representations for images and text. These learned representations are effective for various downstream vision and language tasks. Follow-up works \cite{mu2021slip,zhai2023sigmoid} in this area focus on improving downstream performance by incorporating self-supervision or using other objective functions during pretraining. However, aligning representations at a global (whole image or caption level) is known to only learn coarse-grained features and discard fine-grained visual information. Acknowledging this problem, FILIP \cite{yao2021filip} introduces a cross-modal late interaction mechanism that utilizes a token-wise maximum similarity between image and text tokens to drive the contrastive objective. In the medical domain, GLORIA \cite{9710099} proposes an attention-based framework that uses text tokens to attend to sub-image regions and learns local and global representations. Concurrent to our work, SPARC \cite{bica2024improving} proposes using a sparse similarity metric between image patches and text tokens to learn fine-grained alignment. Our paper shares a motivation to these works in terms of aiming to learn fine-grained representations. However, unlike these methods, we address the fact that image-caption datasets like Conceptual Captions \cite{sharma2018conceptual} or LAION \cite{schuhmann2022laion} contain noisy captions that lack descriptive information, thereby limiting the gains that such fine-grained region-token matching objectives can achieve. Secondly, our approach focuses on learning visual representations that would be able to leverage complementary information at test-time (in the form of LLM-generated descriptions as proposed by \cite{menon2023visual,pratt2023does}
) to recognize fine-grained or novel entities. Finally, in principle, these methods are orthogonal to our contributions and can be coupled with our method.

\vspace{4pt}
\noindent\textbf{Zero-shot Learning with CLIP.}
In image classification, Zero-shot learning methods aim to recognize novel entities that were not seen during training. Relevant to our work, Menon \& Vondrick \cite{menon2023visual} leverage category descriptions generated from a Large Language Model (LLM) as auxiliary information to augment the zero-shot performance of CLIP. On similar lines, CuPL \cite{pratt2023does} and Ren et. al. \cite{ren2023chatgptpowered} use LLMs to generate descriptions in the form of long, cohesive sentences or via nuanced, hierarchy-aware comparisons. TAP \cite{mirza2023tap} learns a text classifier mapping descriptions to categories during training which is used to map from images to categories at test-time. Different from these works, our method aims to improve alignment between images and descriptions that would further bolster the efficacy of using descriptions at test-time.

\vspace{4pt}
\noindent\textbf{Object-aware Vision-Language Pretraining.}
Encouraging object-oriented representations within a vision-language pretraining objective \cite{chen2020uniter,li2020oscar,lu2019vilbert, wang2022objectaware,zhang2023helping,chen2023what,burns2023makes} has been shown to facilitate learning of robust models that can positively impact downstream performance across a variety of tasks in vision-language, video understanding and embodied AI. Many of these approaches follow the DETR line of works \cite{carion2020endtoend,kamath2021mdetr,lei2021qvhighlights} that introduce a query-transformer backbone for detection and grounding. We take inspiration from these works to develop our architecture for encoding visual information. However, our approach only uses the grounding task as an auxiliary objective to distill information from local regions to global representations. We leverage our synthetic descriptions to supervise this grounding module, which is then disabled during evaluation as detailed earlier.

\vspace{4pt}
\noindent\textbf{Universal Visual Recognition.} Recent works \cite{hu2023open,conti2023vocabularyfree} introduce the problem of universal visual recognition or vocabulary-free image classification, where the motivation is to free models like CLIP from a constrained vocabulary thereby allowing classification from an unrestricted set of concepts. Corroborating with our claims, these works observe limitations of CLIP toward recognizing novel examples and fine-grained entities. These works formalize this problem and introduce retrieval-based methods as an initial step towards a solution.

\vspace{4pt}
\noindent \textbf{Multimodal Large Language Models.} MLLMs like LLaVA \cite{liu2024visual}, GPT-4V \cite{gpt4v}, Mini-GPT4 \cite{zhu2023minigpt} integrate image tokens into LLMs, leveraging their powerful reasoning capabilities. MLLMs has been found useful in tasks such as scene understanding \cite{rasheed2023glamm}, story-telling \cite{dong2023dreamllm} etc., where a comprehensive understanding of the images and text is required. We leverage their ability for visual comprehension to generate a set of descriptions for an input image that are used to supervise our fine-grained losses during training. 

\vspace{4pt}
\noindent \textbf{Zero-shot Learning for Images.} Zero-shot learning (ZSL) learning is a challenging problem that requires methods to recognize object categories not seen during training. Various approaches \cite{5206594, 6126281} have proposed using side information like attributes, hierarchical representations etc. to learn a generalizable mapping. More recent efforts \cite{8578679,ye2022zerogen} explore the use of generative models to synthesize useful features for unseen categories. Our method aligns more closely with the former, as we learn a fine-grained correspondence conducive for zero-shot classification by leveraging descriptions as side-information.

\section{Implementation Details}
All baselines reported in the main paper (except LLaVA) utilize a ViT-B/16 architecture as the vision encoder. In Table \ref{table:zeroshot-sup}, we report results using the \textbf{ViT-L/14} architecture trained on CC12M. For encoding text, we utilize a 12-layer transformer network as used with CLIP \cite{radford2021learning}. The outputs from the vision encoder are 768 dimensional, which are then projected to 512. The outputs embeddings obtained from the decoder are also passed through separate projection layers. The projection layer is shared between all region output embeddings and a separate projection layer is used for the image output embedding. Similarly, the text-encoder output is projected to be of the same 512 dimensional size. Additionally, a two-layer MLP with output size 4 is used to regress on the bounding boxes conditioning on the region output embeddings. The supervision for bounding boxes is obtained through the OWLv2 detector, which is originally for a 960 $\times$ 960 resolution image which is down-scaled to 224 $\times$ 224 following the input resolution of our model. While generating these bounding box annotations from OWLv2, we use a confidence threshold value of 0.3. 

\section{Additional Results and Baseline Analysis}

In Table \ref{table:zeroshot-sup}, we report performance of GRAIN and competing baselines when using the \textbf{ViT-L/14} transformer backbone. It shows that our approach is able to consistently outperform all baselines under various Vision Transformer backbones.

\vspace{4pt}
\noindent\textbf{CLIP.} We train the same ViT variant on the same Conceptual Captions (CC3M and CC12M) datasets as our GRAIN model. For performing zero-shot testing on all reported datasets, we use the handcrafted prompts specific to each dataset as introduced in the official code-base \cite{radford2021learning}. These hand-engineered prompts improve the zero-shot performance of CLIP beyond the vanilla, \verb|A photo of {classname}| style prompts.

\vspace{4pt}
\noindent \textbf{CLIP*.} With the introduction of the decoder and bounding-box modules, our method, GRAIN, uses $\sim$ 22\% more parameters compared to CLIP. For a more fair comparison in terms of number of parameters, we report performance for CLIP by using the same architecture as ours, but with the localization modules turned off. We refer to this baseline as CLIP*

\vspace{4pt}
\noindent \textbf{Menon\&Vondrick.} We leverage the official code-base \cite{menon2023visual} to report performance for this baseline. In the main paper, we implement this baseline on top of the CLIP method as per the norm.

\vspace{4pt}
\noindent \textbf{CuPL.} Similarly, we implement the CuPL baseline leveraging official code \cite{pratt2023does} and report performance with CLIP and CLIP* in the main paper and in Table \ref{table:zeroshot-sup} respectively. CuPL shows a similar trend of improving over CLIP baselines but trailing behind our method.

\vspace{4pt}
\noindent \textbf{LLaVA + CLIP.} We use a pretrained LLAVA v1.6 checkpoint from huggingface \cite{wolf2020huggingfaces} that is composed of a ViT-L/14 vision encoder and a Vicuna-13B LLM. The vision and text encoders of LLAVA have been separately pretrained on billion-scale datasets and conjoined through a projection layer. LLaVA has been trained through multiple stages on a specialized set of $\sim$ 150k instructions. Being a generative model, we ask LLAVA to predict a category for an image by using prompts specific to each dataset as described in Table \ref{tab:prompts}. Next, we use a pretrained CLIP text encoder to map the answer generated by LLaVA to the closest category in the vocabulary of the dataset being evaluated on. We use this mapped category as the prediction to compute the top-1 accuracy as usual. We call this method \textbf{LLaVA + CLIP}. Observing Table \ref{table:zeroshot-sup}, our approach is able to reach and even surpass LLaVA's performance on several datasets despite having orders of magnitude smaller parameters and training datasets.
\begin{figure}[t]
    \centering
    \centering
    \includegraphics[width=0.4\textwidth]{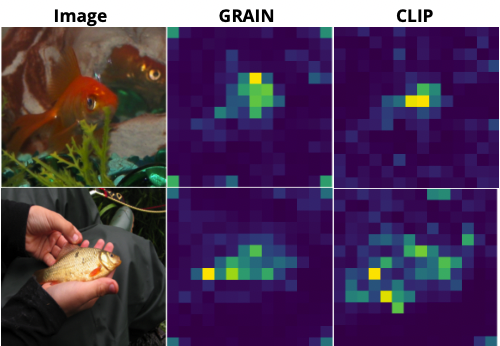}
    \caption{Attention maps show more effective object localization by our model compared to CLIP.}
    \label{fig:attention}
    \vspace{-5mm}
\end{figure}

\vspace{4pt}
\noindent \textbf{FILIP.} Although FILIP \cite{yao2021filip} shares a similar motivation to our method, we note that the approach taken for FILIP is orthogonal to ours. FILIP employs a cross-modal late interaction mechanism to learn associations between image patches and caption tokens without using any side information. In contrast, our approach leverages complementary information in the form of descriptions and their corresponding localizations to learn fine-grained alignments. Being orthogonal to our contributions, in principle, the late interaction mechanism from FILIP can be coupled with our approach. Secondly, a concurrent work \cite{bica2024improving} finds FILIP's results challenging to reproduce due to high training instability, and in practice, observe FILIP to substantially underperform even the zero-shot performance of CLIP on classification tasks. For these reasons, we refrain from comparing our method to FILIP.

\section{Qualitative Analysis}
\noindent \textbf{Visualizing Attention Maps.}
We visualize attention maps of the penultimate encoding layer for our GRAIN model and CLIP in Figure \ref{fig:attention}. Our model is seen to effectively focus on the object regions in the image which stems from our localization and alignment objectives.

\vspace{4pt}
\noindent \textbf{Recognizing Novel Classes.}
In this experiment, we focus on recognizing newly popular entities, namely the \verb|Apple Vision Pro| and \verb|Tesla Cybertruck|, which emerged after datasets like Conceptual Captions were constructed. First, we add these two classnames to the Imagenet-1K  vocabulary. Next, to simulate a real-world open-vocabulary scenario, we also include three related but distinct categories for each novel entity, making this a challenging task. Specifically, for the \verb|Apple Vision Pro|, we add competing Virtual Reality (VR) headsets such as the \verb|Meta Quest 2|, \verb|Microsoft Hololens|, and \verb|Google Glass|. For the \verb|Tesla Cybertruck|, we include other pickup trucks like the \verb|Rivian R1T|, \verb|Ford F-150|, and \verb|Toyota Tundra|.
We then utilize GPT-3 (language only) to generate descriptions for the concepts in this extended vocabulary. Following the inference-time procedure discussed in the main paper, we present the top-5 predictions made by both our model and \cite{menon2023visual} in Figure \ref{fig:topk}. Our findings indicate that our model consistently identifies the correct class names with high confidence, whereas the baseline is able to include it in top-5 but fails to rank them as the top choice. This highlights our models ability to recognize novel concepts by leveraging the learned image-description alignment. 

\vspace{4pt}
\noindent \textbf{Description Grounding.}
To showcase the efficacy of our grounding module, we present visualizations of its predictions in Figure \ref{fig:grounding}, with images from the Imagenet dataset. We include additional visualizations in the Appendix.  These visualizations include LLM-generated descriptions and the corresponding bounding boxes predicted by our model, with each matched pair coded by color. We also include descriptions belonging to this class that are not matched to a bounding box.
\begin{figure}[t]
    \centering
    \centering
    \includegraphics[width=0.5\textwidth,height=0.25\textwidth]{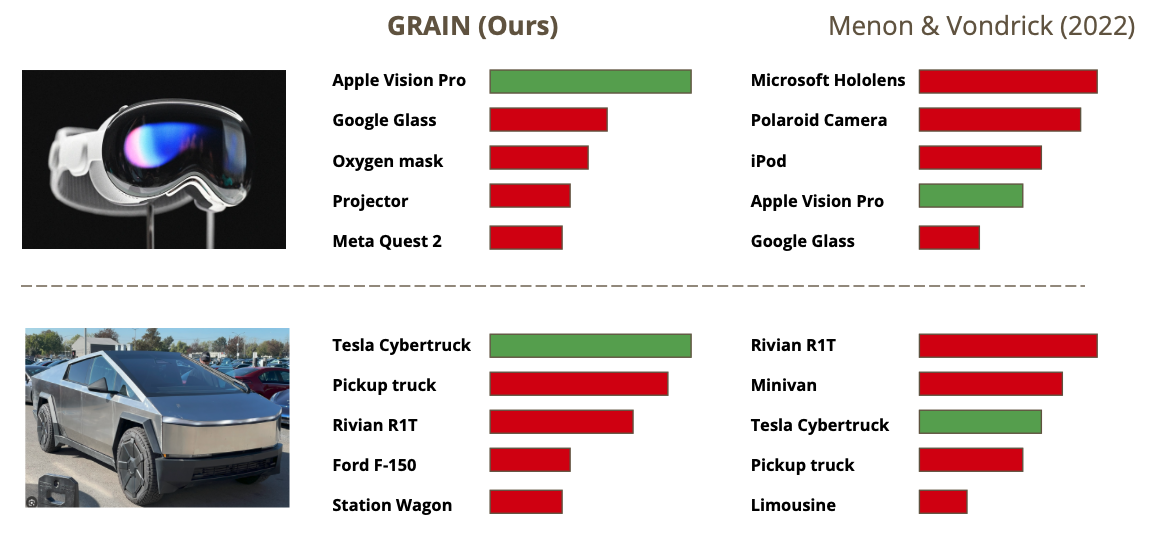}
    \caption{Visualization of top-5 predictions of our model on novel entities alongside \cite{menon2023visual}. Our method consistently identifies the ground truth class as the top prediction.}
    \label{fig:topk}
\end{figure}

\begin{figure*}[t]
    \centering
    \centering
    \includegraphics[width=0.95\textwidth]{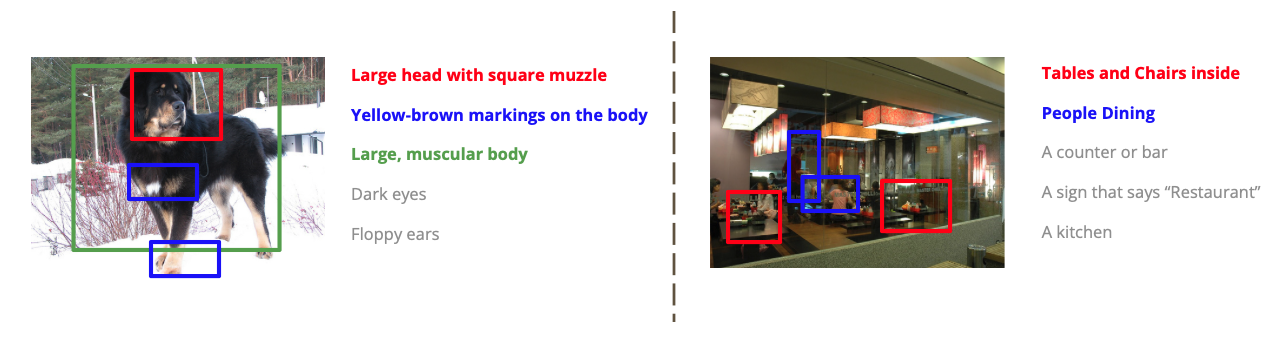}
    \caption{Localization and region-description matching predictions made by our model on images from ImageNet.}
    \label{fig:grounding}
\end{figure*}

\section{Pretraining Dataset Details}
We train all models on the CC3M and CC12M datasets. As explained earlier, to obtain description and localization annotations, we prompt LLaVA in two stages to extract the primary visual subject of the image and then gather image descriptions by asking LLaVA to focus on the identified visual subject. We obtain bounding boxes corresponding to each description by using OWLv2 \cite{minderer2024scaling}, an off-the-shelf open-vocabulary object detector. We filter the predicted boxes using a confidence threshold of 0.3 to discard noisy predictions. 
\begin{figure*}[t!]
    \centering
    \includegraphics[width=0.87\textwidth]{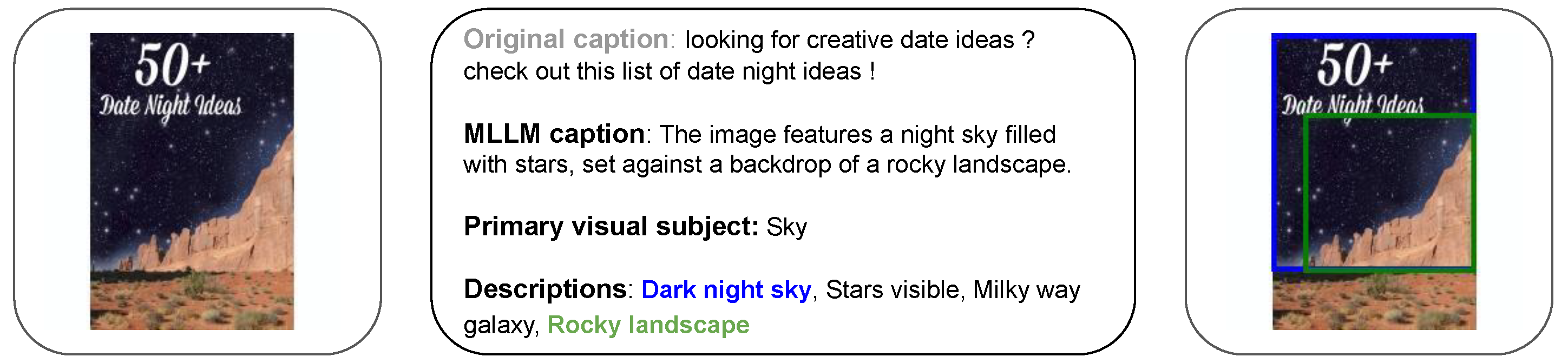}
    \includegraphics[width=0.87\textwidth]{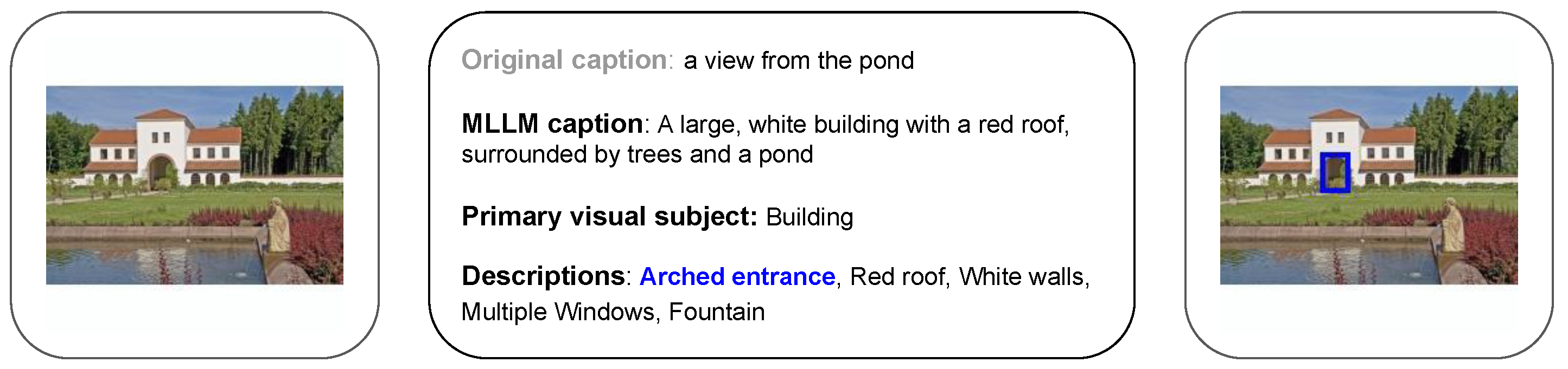}
    \includegraphics[width=0.87\textwidth]{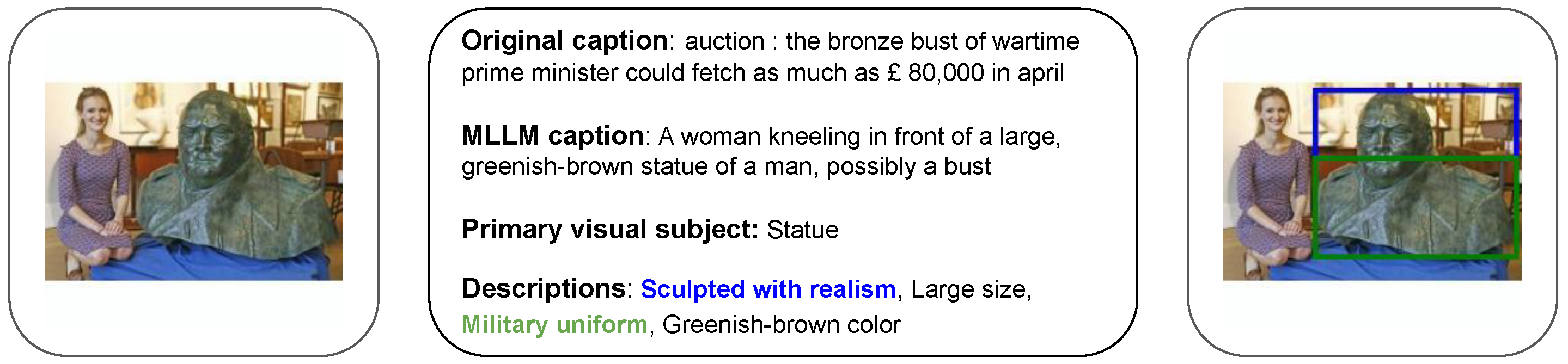}
    \includegraphics[width=0.87\textwidth]{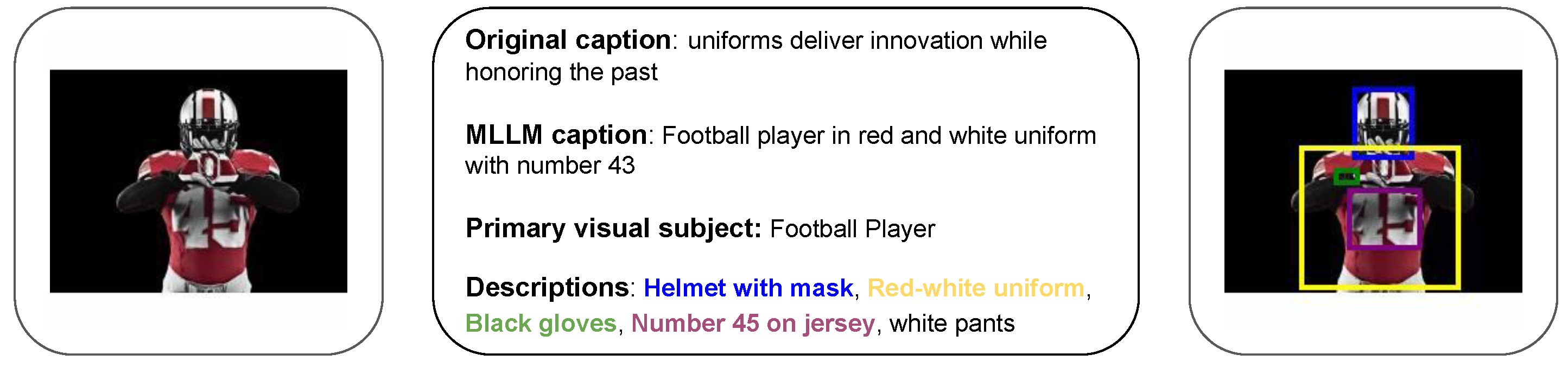}
    \includegraphics[width=0.87\textwidth]{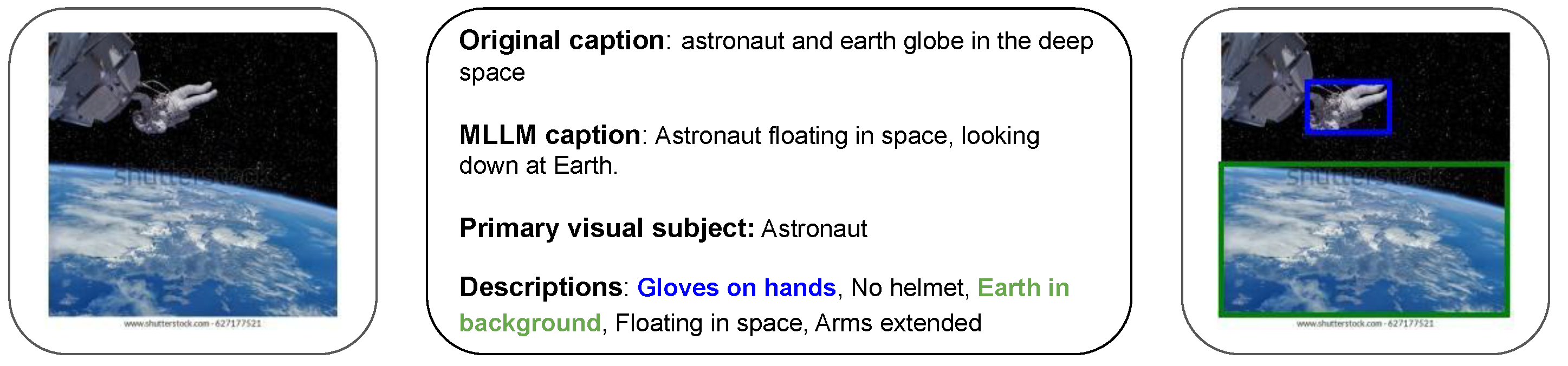}
    \includegraphics[width=0.87\textwidth]{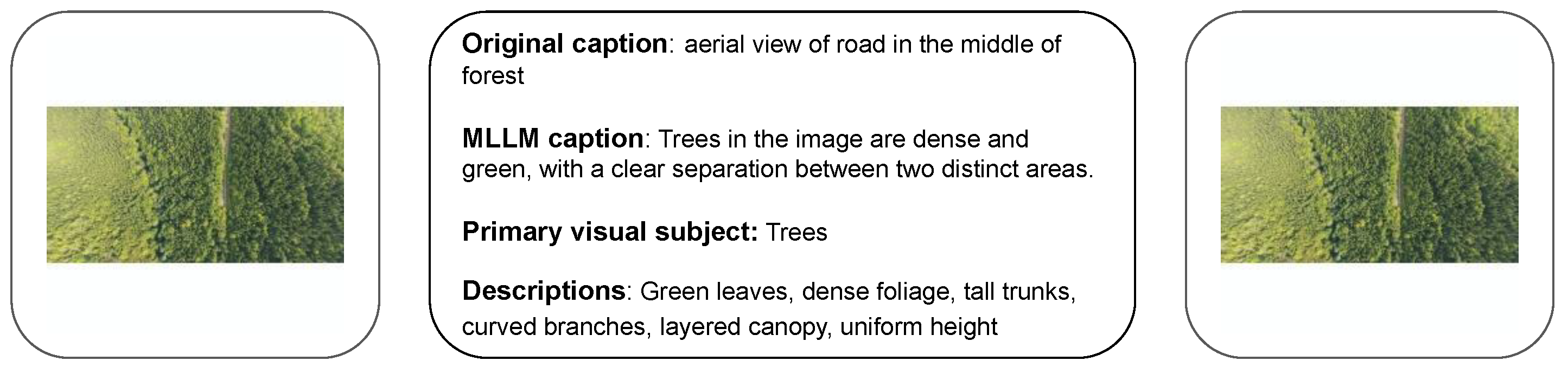}
    \caption{Sample annotations generated using our two-stage LLaVA prompting scheme followed by OWLv2 localization.}
    \label{fig:tryer}
\end{figure*}

\section{Products-2023 Dataset Details}
To evaluate our approach's ability to recognize novel samples, we manually curated a dataset comprising 1,500 images spanning 27 distinct categories. These images were carefully filtered and labeled through a manual process. Specifically, we compiled a list of products launched after 2023, scraped corresponding images, and performed manual filtering and labeling. Since the pretraining datasets used in our setup (CC3M, CC12M) were finalized prior to 2023, this dataset represents novel examples. The full list of categories includes:  [\textit{Apple Vision Pro}, \textit{Grimace Shake(drink from McD)}, \textit{Starry (drink from Pepsi)}, \textit{Playstation Portal}, \textit{Apple Watch Ultra 2}, \textit{Apple Watch Series 9}, \textit{Samsung Galaxy Watch 6}, \textit{Xiaomi Smart Band 8}, \textit{Kia K4}, \textit{Rivian R2}, \textit{Honda Ye GT}, \textit{Ferrari 12Cilindri}, \textit{Renault 5 E-Tech}, \textit{Toyota Tundra}, \textit{Ford F-150 Lightning}, \textit{Tesla Cybertruck}, \textit{Xiaomi SU7}, \textit{Lamborghini Revuelto}, \textit{Hyundai Mufasa}, \textit{Wordle}, \textit{Asus ROG Ally}, \textit{Meta Quest Pro}, \textit{Microsoft Hololens}, \textit{Google Glass}, \textit{Prime (drink)}, \textit{iPhone 15}, \textit{Google Pixel 8}
]. We intend to release this dataset with the final version of this paper.

\section{Sample Annotations}
In Figure \ref{fig:tryer}, we illustrate the annotations obtained using our two-stage LLaVA prompting followed by bounding box prediction using OWLv2. We randomly select images and captions (original caption) from the CC3M dataset and present the corresponding MLLM caption, primary visual subject, and descriptions generated by our annotation pipeline. The descriptions are color-coded by their associated bounding box. Overall, our annotation pipeline is effective in identifying the primary visual subject, which is the most prominent object or concept in the image, and generating descriptions and corresponding localizations by focusing on this subject. The first five rows show cases where the pipeline successfully localized at least one description, whereas the last row demonstrates a case where no description could be localized due to the vague nature of the image, making the descriptions difficult to localize.

\begin{table*}[t]
\centering
\caption{Prompts to LLaVA for the zero-shot visual recognition task in Table \ref{table:zeroshot-sup}.}
\label{tab:prompts}
\resizebox{\textwidth}{!}{%
\begin{tabular}{L{0.2\textwidth} L{0.7\textwidth}}
\toprule
\rowcolor{white} \textbf{Dataset} & \textbf{Prompt} \\
\midrule
\rowcolor{rowgray} DTD             & Fill in the blank: this is a photo of a \{\} texture \\
Pets                        & What animal is in the image? Be specific about the breed. Fill in the blank: this is a photo of a \{\} \\
\rowcolor{rowgray} Places365        & What place is this in the image? Fill in the blank: this is a photo of a \{\} \\
Food101                            & What food is in the image? Fill in the blank: this is a photo of a \{\} \\
\rowcolor{rowgray} Cars            & What type of car is in the image? Be specific about the make and year. Fill in the blank: this is a photo of a \{\} \\
Others                             & Fill in the blank: this is a photo of a \{\} \\
\bottomrule
\end{tabular}%
}

\end{table*}
\section{Two-stage versus Single stage Annotation}
In this work, we employ a two-stage annotation pipeline to elicit descriptions from LLaVA. Specifically, in the first stage, we prompt LLaVA to identify the primary visual subject in the image, followed by generating descriptions for this subject. We observe that this approach leads to the generation of descriptions that are more specific and focused on the constituent regions in the image that make up the subject. In Figure \ref{fig:compare}, we compare the descriptions generated by this strategy with a single-stage pipeline that directly prompts LLaVA to generate descriptions without first identifying the subject. We randonly pick samples from the CC12M dataset to illustrate the difference. Contrasting the two setups, we can see that the two-stage approach produces more specific descriptions that are well-grounded in the image compared to the one-stage approach, which either outputs overly generic descriptions or tends to hallucinate (See Rows 1 and 2). This issue is more pronounced for complex scenes involving unusual or fine-grained objects.

\section{Failure modes in our grounding module}
In the main paper, we showcased examples where the grounding module of our approach successfully localized descriptions in images. In this section, we particularly highlight failure cases where the model is unable to correctly localize descriptions within the image. Following the same setup as the main paper, we use images from ImageNet and descriptions generated from an LLM by following Menon \& Vondrick's \cite{menon2023visual} strategy of prompting GPT-3 (language-only) with category names. It is important to note that since these descriptions were generated using only the category name and without access to images, some descriptions might not be visible in every image. We expect our approach to localize descriptions that are present in an image and not localize those that are absent. While our approach effectively grounds descriptions on average, we illustrate failure cases in Figure \ref{fig:localize}.

Row 1 includes partially successful cases, where the model localizes descriptions but the bounding boxes are either slightly off the mark or does not localize all instances of that description in the image.

Row 2 includes examples where either the model cannot localize a single description in the image or incorrectly associates the description with another region in the image. (the description \verb|typically orange or brown| refers to the \textit{basketball} but was incorrectly assigned to the \textit{jersey of the player} that has a similar color.)

Row 3 includes cases of hallucination, where the model localizes descriptions that are not present in the image.

\begin{figure*}[t!]
    \centering
    \includegraphics[width=0.87\textwidth]{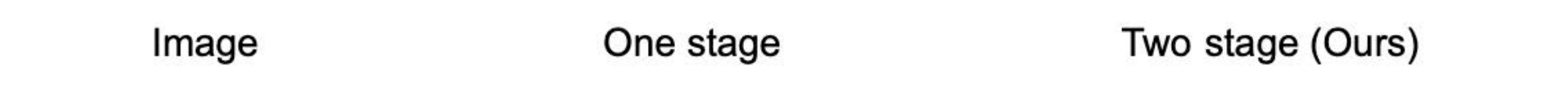}
    \includegraphics[width=0.87\textwidth]{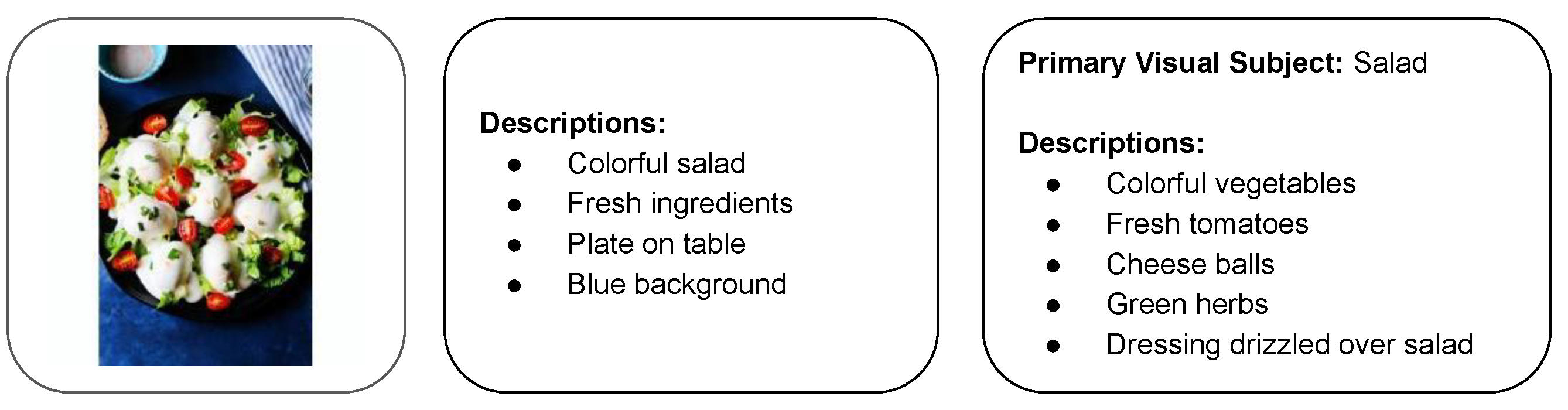}
    \includegraphics[width=0.87\textwidth]{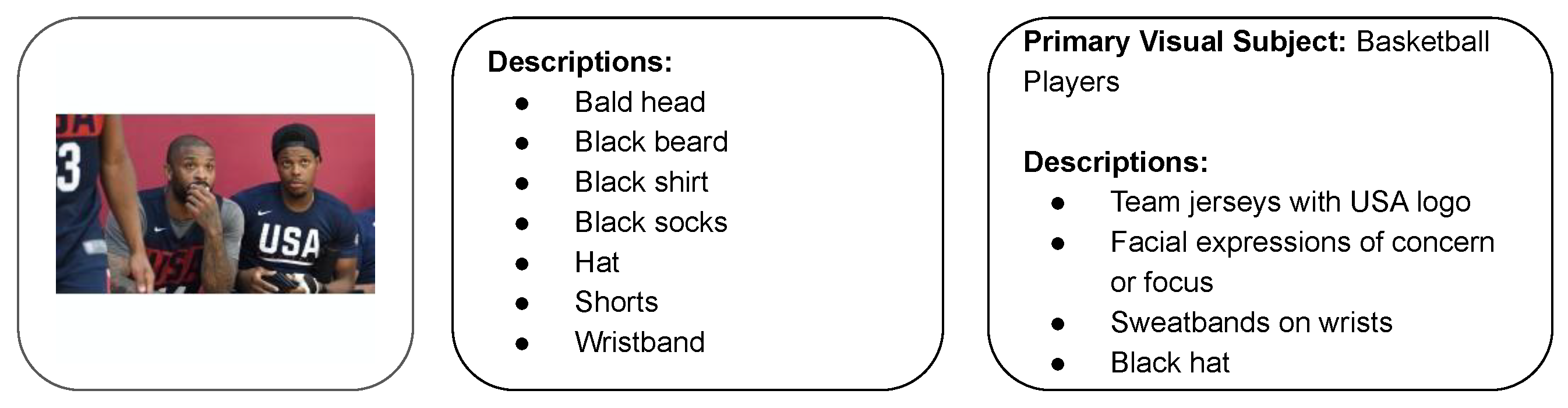}
    \includegraphics[width=0.87\textwidth]{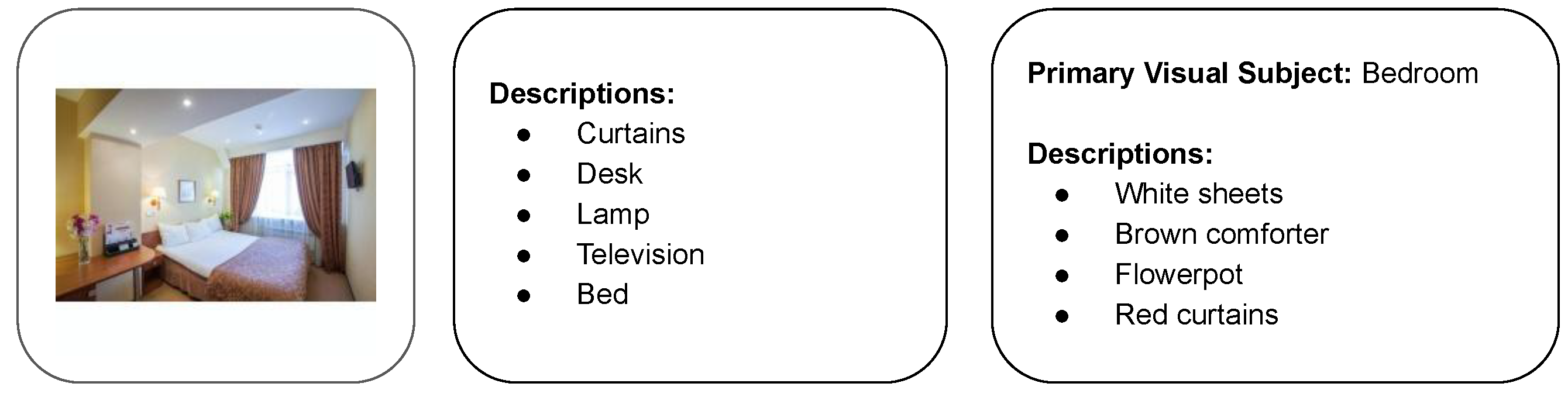}
    \includegraphics[width=0.87\textwidth]{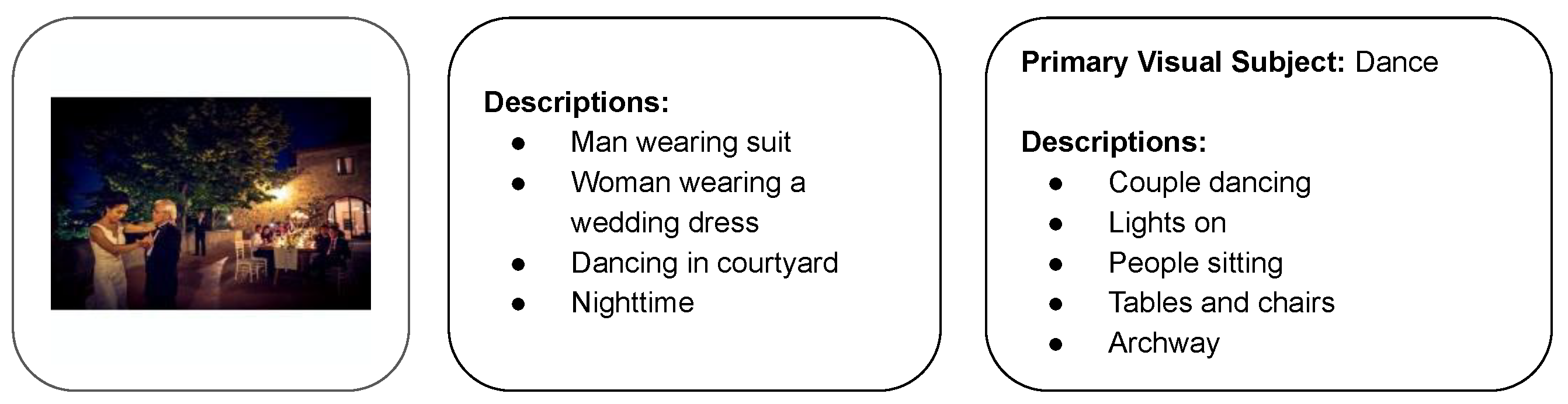}
    \includegraphics[width=0.87\textwidth]{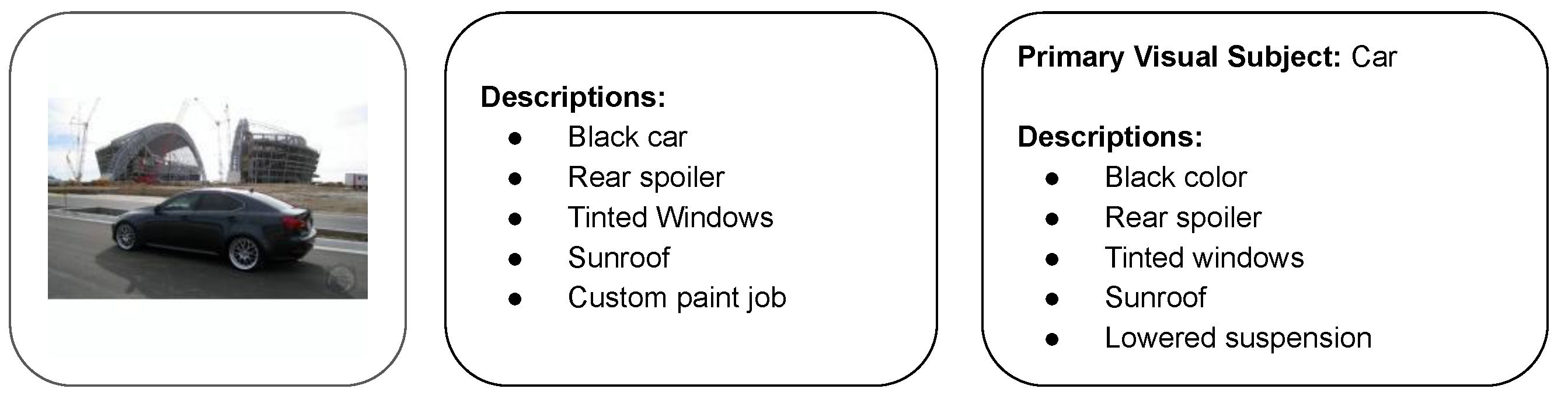}
    \caption{Qualitative comparison between one-stage (middle) and two-stage (right) LLaVA-based annotation schemes.}
    \label{fig:compare}
    \vspace{1cm}
\end{figure*}
\begin{figure*}[t!]
    \centering
    \includegraphics[width=0.87\textwidth]{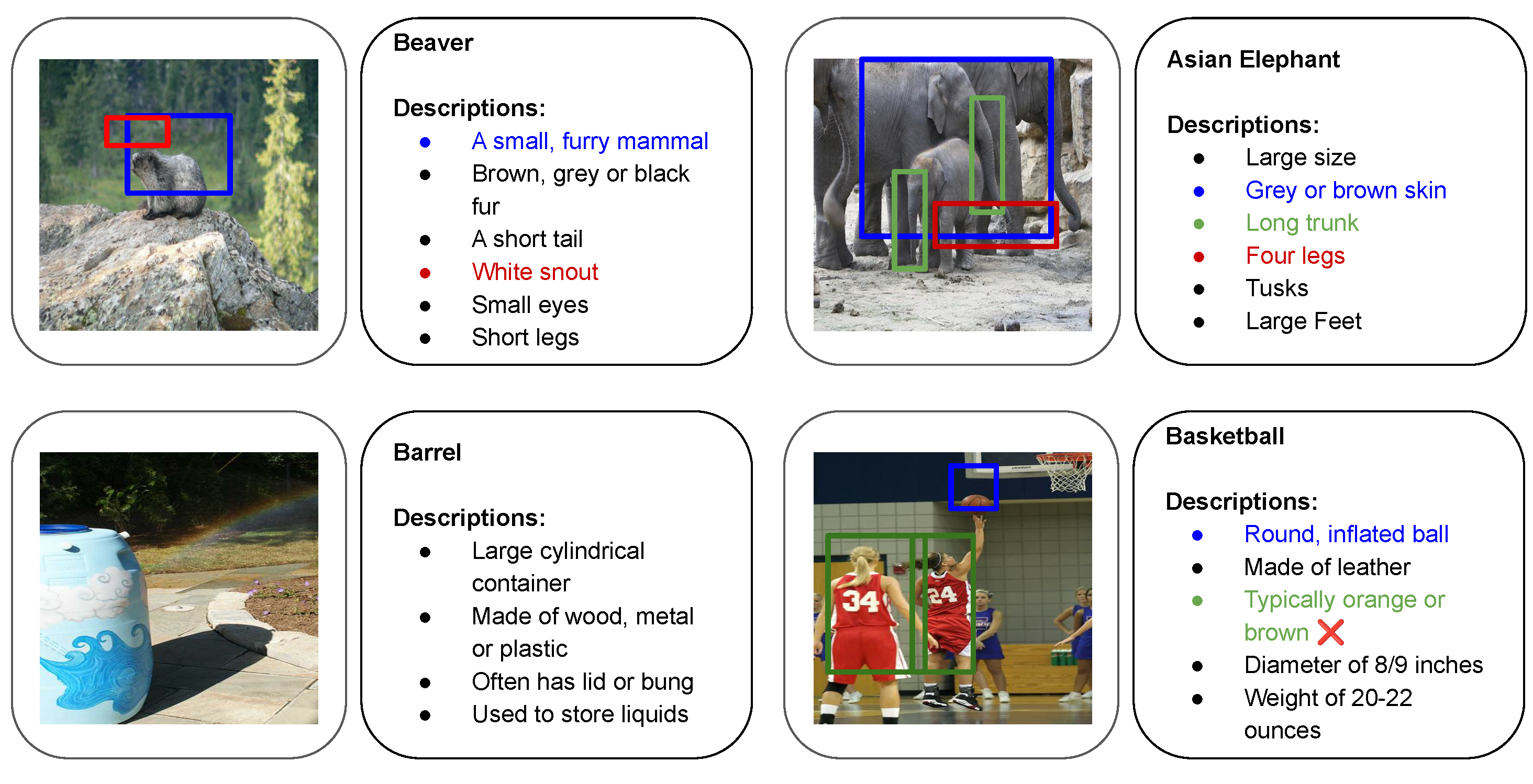}
    \includegraphics[width=0.87\textwidth]{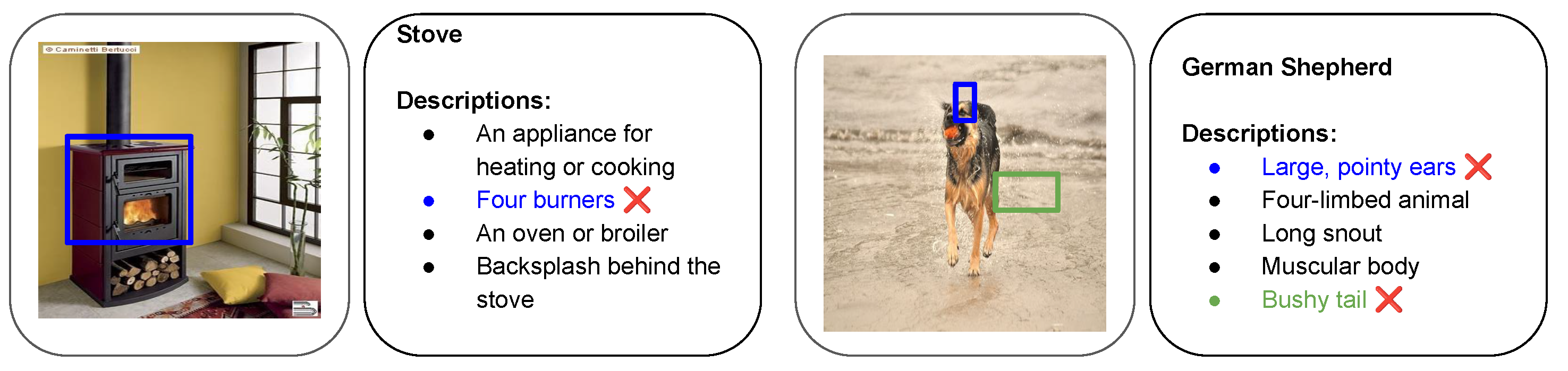}
    \caption{Visualization of failure modes from our grounding module on ImageNet-1K.}
    \label{fig:localize}
\end{figure*}

\section{Limitations and Broader Impact}

\vspace{4pt}
\noindent\textbf{Limitations.}  Our method achieves substantial gains over CLIP and other baselines on zero-shot transfer tasks such as image classification, attribute-based image classification, and cross-modal retrieval. These improvements can be attributed to the fine-grained region-to-description associations learned by our model during the training process. However, learning these correspondences requires annotations in the form of descriptions and bounding box localizations, which are computationally expensive to obtain. As mentioned earlier, our annotation scheme demands significant GPU resources and can take long hours for large datasets. Additionally, since we do not filter or curate these annotations, it might result in some misaligned or inaccurate descriptions or captions, which might not provide the correct signal during the learning process. Future work could explore the use of efficient models to generate annotations as well as a filtering mechanism to ensure all generated text and bounding boxes are correctly aligned with the semantic content of the image.

\vspace{4pt}
\noindent\textbf{Broader Impact.}  We propose a strategy to learn fine-grained image-text correspondences without requiring additional human annotations. Our approach leverages weak supervision from Multimodal Large Language Models (MLLMs) to train a region-aware model that strongly outperforms CLIP across several tasks and datasets. Despite having significantly smaller parameters and training costs, our approach matches and sometimes even outperforms LLaVA, a state-of-the-art MLLM, on zero-shot visual recognition. Although obtaining these annotations is computationally expensive, once acquired, our approach can be viewed as enabling the training of smaller models with small-scale datasets to achieve performance equivalent to a large model trained on extensive data, potentially making Vision and Language Model (VLM) training more accessible. Further integration of our approach with retrieval-based systems and multimodal LLMs is an interesting future direction.

\end{document}